\documentclass[10pt,twocolumn,letterpaper]{article}

\usepackage{cvpr}

\def\ours{STAT}

\definecolor{cvprblue}{rgb}{0.21,0.49,0.74}
\usepackage[pagebackref,breaklinks,colorlinks,allcolors=cvprblue]{hyperref}
\usepackage{multirow}   
\usepackage{mwe}
\usepackage{cuted}
\usepackage{arydshln}

\title{Soft Tail-dropping for Adaptive Visual Tokenization}

\author{
Zeyuan Chen$^1$ \quad
Kai Zhang$^2$ \quad
Zhuowen Tu$^1$ \quad
Yuanjun Xiong$^2$ \quad
\\
$^1$UC San Diego \quad $^2$Adobe
}

\begin{document}
\maketitle

\begin{strip}
    \centering
    \includegraphics[width=1.0\linewidth]{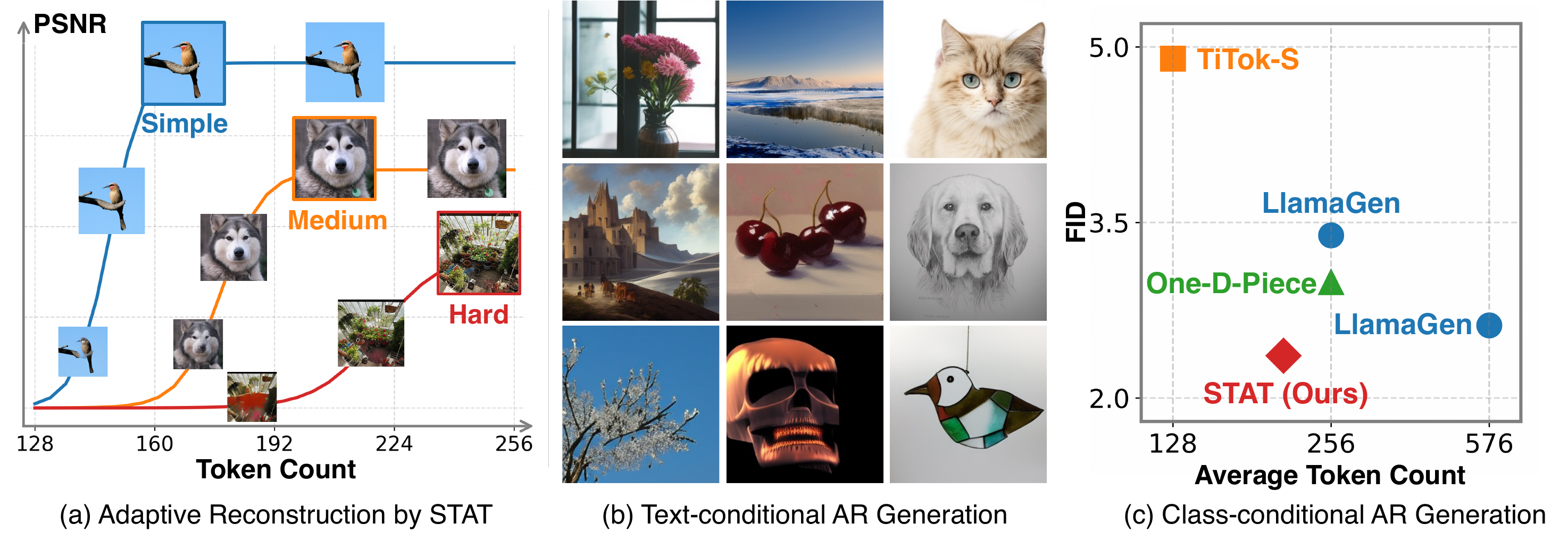}
        \captionof{figure}{\textbf{(a) \ours{} adaptively allocates token counts based on image complexity} (the outlined figures), using fewer tokens for simple images and more for complex ones. When paired with a vanilla autoregressive model, \ours{} enables \textbf{(b) high-quality text-conditional image generation} and \textbf{(c) the best FID in class-conditional generation} on ImageNet among visual tokenizers.}
    \label{fig:teaser}
\end{strip}

\begin{abstract}
We present \textbf{S}oft \textbf{T}ail-dropping \textbf{A}daptive \textbf{T}okenizer (STAT), 
a 1D discrete visual tokenizer that adaptively chooses the number of output tokens per image according to its structural complexity and level of detail. STAT encodes an image into a sequence of discrete codes together with per-token keep probabilities. Beyond standard autoencoder objectives, we regularize these keep probabilities to be monotonically decreasing along the sequence and explicitly align their distribution with an image-level complexity measure. As a result, STAT produces length-adaptive 1D visual tokens that are naturally compatible with causal 1D autoregressive (AR) visual generative models. On ImageNet-1k, equipping vanilla causal AR models with STAT yields competitive or superior visual generation quality compared to other probabilistic model families, while also exhibiting favorable scaling behavior that has been elusive in prior vanilla AR visual generation attempts.
\end{abstract}    
\section{Introduction}
\label{sec:intro}
Autoregressive models have demonstrated remarkable success across modalities - powering large language models (LLMs)~\cite{Radford2018GPT,touvron2023llama} in text and achieving strong performance in visual generation for both images and videos~\cite{chen20iGPT}, as well as joint modeling of language, vision, and beyond~\cite{cui2025emu35nativemultimodalmodels,deng2025bagel,chen2025cvp}. Being the cornerstone of this success, modern text tokenizers~\cite{sennrich2016BPE,radford2019language} demonstrate the intriguing property that its representation cardinality is dependent on the content it is representating. This property of content-dependent compression, is unsurprisingly also the hallmark of modern image and visual compression approaches. However, tokenizers for popular visual generation models~\cite{sun2024autoregressive,peebles2023scalable,chang2022maskgit} still anchor the visual tokens on rigid 2D structure and enforce fixed compression ratio regardless of the image and video content. Consequently, simple images may be over-represented, while complex scenes may remain under-encoded, which in turn challenges the capability of the generative models. Despite numerous efforts~\cite{yu2024image, kim2025democratizing,miwa2025one,bachmann2025flextok,rippel2014learning} to decouple the representation cardinality of visual tokenizer from the input resolution, we have not achieved adaptive alignment of output token cardinality with the visual complexity. We also have not observed empirically the compatibility of content-adaptive visual tokenizers with generative models, especially autoregressive generative models, which is well suited for tokenizers with this property~\cite{Radford2018GPT,chen20iGPT}.

In this work, we propose \ours{}, which uses soft tail-dropping for adaptive visual tokenization. In detail, as a tokenizer, \ours{} is able to encode a given image into discrete latent tokens, decide the token counts used for a given image, and use the corresponding token counts for decoding. Concretely, \ours{} introduces a probabilistic token dropout module that predicts keep probabilities for each token position, followed by per-token Bernoulli sampling to decide whether a token is kept or dropped. To guide the learning process for the dropout probabilities, we apply three priors: (1) the probability profile should be monotonically decreasing, enforcing a tail-shaped dropping pattern; (2) the sum of predicted probabilities should correlate with image complexity, assigning fewer tokens to simple images and more to complex ones; and (3) the model should attempt to use fewer tokens to increase the compression ratio. These priors induce a model providing adaptive compression over images, which is demonstrated to have fewer average token count but better reconstruction.

More importantly, the adaptivity of \ours{} facilitates ease of modeling for generative models. When integrated with a vanilla autoregressive model that performs next token prediction, \ours{} enables the model to reach state-of-the-art generation quality, comparable to or surpassing diffusion-based models and other autoregressive models deviating from the simple next token prediction paradigm.

To summarize, our contributions are:
\begin{itemize}
    \item We propose \ours{}, an adaptive visual tokenizer that dynamically determines the compression ratio with different token counts based on image complexity.
    \item We introduce to learn a probabilistic soft tail-dropping strategy that enables the tokenizer to adaptively allocate token counts to different images, leading to improved robustness in generative modeling.
    \item \ours{} achieves the state-of-the-art reconstruction performance with fewer average tokens used. Integrating \ours{} into a vanilla autoregressive framework yields a strong generative model comparable with other state-of-the-art generative models.
\end{itemize}

\section{Related Work}
\label{sec:Relatedwork}

\subsection{Image Tokenization}
Image tokenization is one important step in modern visual generative models~\cite{rombach2022high,lee2022autoregressive,chang2022maskgit}, where the images are mapped from the pixel space into a compressed high-dimensional latent space, either continuous or discrete, to avoid prohibitive computation in pixel-space and smoothen the distribution to be modeled. Variational Autoencoder (VAE)~\cite{kingma2013auto} is one representative continuous tokenizer that maps images into a continuous latent space. Vector-quantized autoencoder (VQ-VAE)~\cite{van2017neural} pioneers discrete visual tokenization. It consists of an encoder, a quantizer with a codebook at the bottleneck, and a decoder. The quantizer maps the encoded continuous embeddings into codebook embeddings by nearest neighbor search, and the discretized embeddings are fed to the decoder for reconstruction. Improvements to VQ-VAEs have been widely explored, including architecture~\cite{razavi2019generating,yu2021vector,lee2022autoregressive,chen2025x}, training objectives~\cite{ramesh2021zero,esser2021taming}, and quantization techniques~\cite{el2022image,mentzer2023finite,yu2023language,zhao2024bsqvit}.

VQ tokenizers typically maintain a 2D spatial structure, where the downsampled 2D features at the bottleneck are discretized. To decouple the token number from input dimension, 1D tokenizers like TiTok~\cite{yu2024image,kim2025democratizing} propose to concatenate 2D patch embeddings with 1D latent tokens as the model input and distill information from patch embeddings into 1D latent tokens, which are then quantized. Flexible tokenizers~\cite{yan2024elastictok,bachmann2025flextok,miwa2025one} are subsequently developed to learn ordered 1D token representation with nested dropout~\cite{rippel2014learning}, allowing representing images with a varying number of tokens. These methods still employs a predetermined token count in their inference. \ours{} learns to reconstruct images with content-adaptive token counts, representing an image with the number of tokens aligned with its complexity. 

Among emerging works investigating adaptive tokenizers, ALIT~\cite{duggal2024adaptive} proposes to iteratively update the 1D token sequences to identify smallest number of tokens for a given image, but it requires multiple forward passes. One concurrent work~\cite{duggal2025single} uses post-hoc heuristics based on reconstruction of the full-length tokens to determine the token count after the encoding. In contrast, \ours{} is single-pass and aligns token cardinality prediction with image complexity via an explicit objective, rather than treating token count as a fixed hyperparameter or a heuristic byproduct. We also demonstrate that \ours{} is particularly suitable for vanilla autoregressive models, achieving compelling generation quality that scales favorably with the capacity of generative models.

\begin{figure*}[t!]
\centering
\includegraphics[width=0.98\textwidth]{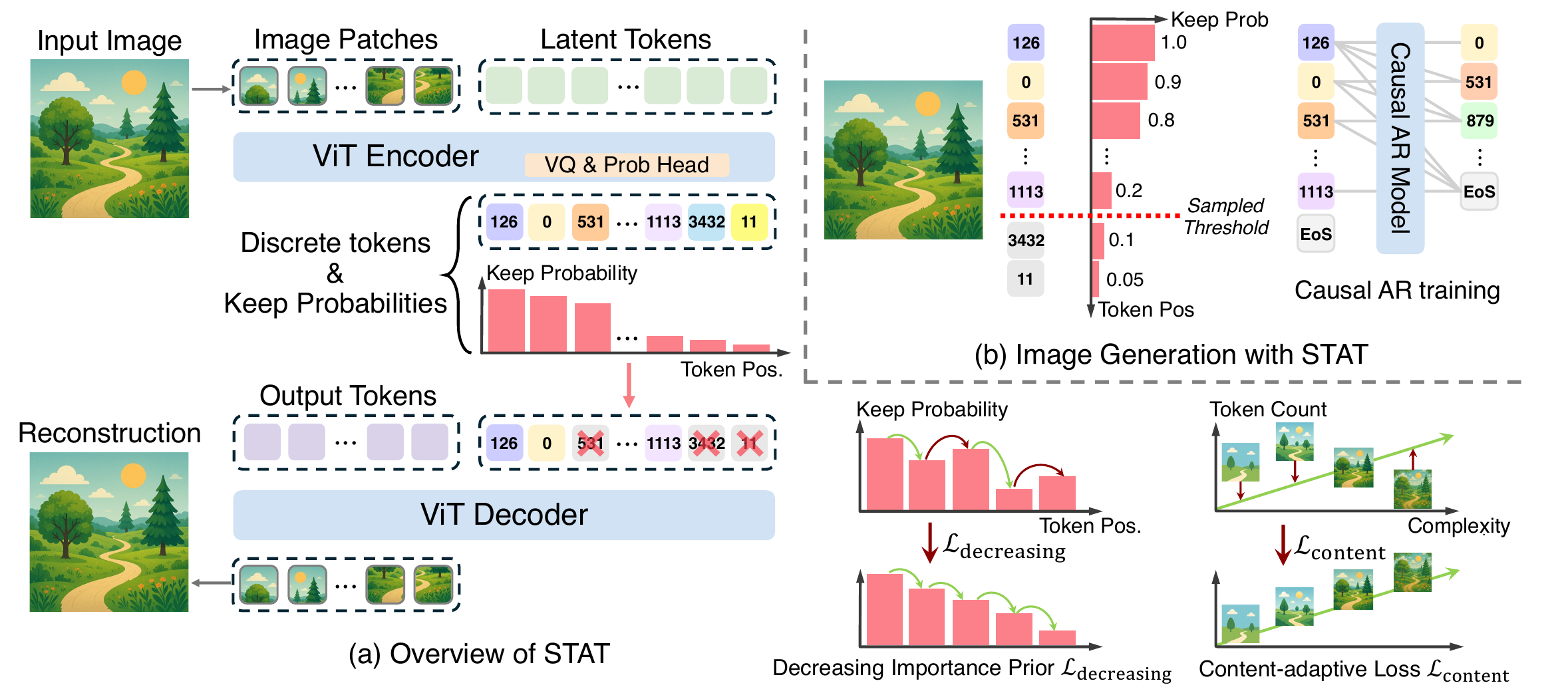}
\caption{\textbf{(a) Overview of \ours{}.} A ViT encoder followed by a vector-quantizer (VQ) layer produces discrete latent tokens along with a keep probability for each token position. Probabilistic token dropping is applied to obtain a masked latent sequence for reconstruction, while two regularization losses enforce content-adaptive allocation and a decreasing importance prior. \textbf{(b) Image generation with \ours{}.} A special End-of-Sequence (EoS) token enables adaptive-length autoregressive generation, where the EoS position is determined by a threshold  with the keep-probability profile predicted by \ours{}.}
\end{figure*}

\subsection{Image Generation}
Generative adversarial network (GAN)~\cite{goodfellow2014generative} is one representative framework for early image geneartion. Many subsequent works~\cite{brock2018large,sauer2022stylegan,kang2023scaling} have been proposed to advance GANs in both fidelity and diversity. More recently, diffusion models~\cite{ho2020denoising,song2021ddim,rombach2022high,peebles2023scalable,Chang_2025_CVPR,mao2025sir,zeng2025yolo} have become a popular choice for image generation due to their excellent generation quality and stable training.

With the development of VQ-VAE and other discrete tokenizers, discrete generative models have also gained attention in recent years, which can be mostly divided into two categories. BERT-style models~\cite{chang2022maskgit,lezama2022improved,yu2023language} that typically use an Transformer encoder architecture with bi-directional attention and sample tokens in random order to gradually unmask all tokens in generation. Another category, GPT-style autoregressive generative models~\cite{esser2021taming,yu2021vector,lee2022autoregressive,sun2024autoregressive} are based on decoder-only architectures with casual attention. It features easy integreation with other modalities such as lanaguage. LlamaGen~\cite{sun2024autoregressive} is one representative work in this category, which demonstrates that with powerful image tokenizers and scalable model architectures, autoregressive model can reach competitive performance with diffusion models, albeit still hitting scaling bottleneck with respect to the model capacity.

Recently there have been numerous efforts in exploring new autoregressive-like pipelines for visual generation. VAR~\cite{tian2024visual} models the autoregressive learning on images as next-scale prediction to form a coarse-to-fine generation process. MAR~\cite{li2024autoregressive} introduces a diffusion loss to BERT-style generative models to allow using continuous tokenizers. RandAR~\cite{pang2025randar} and xAR~\cite{ren2025beyond} augment next token prediction with different variations to strengthen decoder-only autoregressive visual generation. We note that these methods, while achieving strong generation performance, require additional effort~\cite{Tian2025Infinity,Zhou2024TransfusionPT,deng2025bagel} to unite with LLMs.

\section{Method}

Our goal is to build a visual tokenizer that allocates \emph{content-adaptive}
numbers of latent tokens. This requires moving beyond fixed 2D image grids and learning
a 1D latent sequences whose lengths are determined adaptively by the input complexity. \ours{} addresses
this through a two-stage training pipeline. First, the tokenizer learns prefix-robust
reconstruction via randomized tail truncation, producing an ordered latent
sequence. Second, it becomes content-adaptive by predicting per-token keep
probabilities and learning a soft tail-dropping policy driven by visual
complexity. This enables \ours{} to assign more tokens to complex images and fewer
to simple ones, and naturally supports 1D auto-regressive (AR) visual generation.

\subsection{Architecture}

\ours{} adopts a 1D Transformer-based tokenizer that decouples tokens from fixed 2D grids, enabling decoding with variable-length latents and adaptive token allocation.

Given an image $x \in \mathbb{R}^{B \times C \times H \times W}$, a patch
embedding layer maps it to $\mathbf{P} \in \mathbb{R}^{B \times N \times D}$, where
$N = H/f \cdot W/f$ for downsampling factor $f$.  
We concatenate $\mathbf{P}$ with $L$ learnable latent tokens 
$\mathbf{L} \in \mathbb{R}^{B \times L \times D}$ and encode the full sequence:
\begin{equation}
[z_p, z_l] = \mathrm{Enc}([\mathbf{P},\, \mathbf{L}]).
\end{equation}
where $z_p$ and $z_l$ are output patch tokens and latent tokens.
We only retain $z_l$ and the vector-quantizer maps each latent token from $z_l$ to the nearest codebook entry, yielding discrete latents $z_q$.  
The decoder reconstructs the image from $z_q$ and a set of learnable output
tokens $\mathbf{O}$:
\begin{equation}
\hat{x} = \mathrm{Dec}([z_q,\, \mathbf{O}]).
\end{equation}

This architecture distills any input image into a 1D token sequence,
making \ours{} compatible with both flexible prefix reconstruction
(Sec.~\ref{sec:flexible_tokens}) and content-adaptive token allocation
(Sec.~\ref{sec:adaptive_tokens}).

\subsection{Adaptive Visual Tokenizer}

\ours{} is trained in two stages.  
In the first stage (Sec.~\ref{sec:flexible_tokens}), the tokenizer learns to 
reconstruct images from \emph{flexible} prefix lengths, but the selection 
of token count is externally randomized and not conditioned on content.  
In the second stage (Sec.~\ref{sec:adaptive_tokens}), \ours{} becomes 
\emph{content-adaptive}, learning to allocate token counts based on image 
complexity through a probabilistic soft tail-dropping mechanism.

\subsubsection{Flexible Prefix Reconstruction}
\label{sec:flexible_tokens}

Before learning content adaptivity, we first train the tokenizer to reconstruct images from \emph{any} prefix of its 1D latent sequence.  
Following prior work~\cite{bachmann2025flextok,miwa2025one}, \ours{} is 
trained with \textit{hard tail-dropping}: in each iteration, a keep length 
$K \sim \mathcal{U}(L_\text{min}, L_\text{max})$ is sampled from a uniform distribution and the first $K$ 
quantized tokens are used for reconstruction:
\[
\hat{x} = \mathrm{Dec}([z_q \odot m,\, \mathbf{O}]), \quad 
m_i = \mathbf{1}[i < K].
\]

Random prefix truncation induces an \emph{ordered} token representation in which early tokens encode coarse global structure, while later tokens refine details.This stage enables \ours{} to reconstruct images from prefixes of latent token sequences, but does \emph{not} teach the tokenizer to determine how many tokens each image should use.

\subsubsection{Content-Adaptive Token Allocation}
\label{sec:adaptive_tokens}

To make token counts adaptive to visual complexity, \ours{} learns to predict a per-token keep-probability profile for each image and applies probabilistic soft tail-dropping.

\noindent \textbf{Per-token Keep Probabilities.} We design our encoder to output latent tokens $z_l \in \mathbb{R}^{B \times L \times D}$ along with their per-token keep probability by:
\begin{equation}
    p_{j,i} = \sigma\!\big(g_\theta(z_l[j,i])\big).
\end{equation}
where $j$ indexes the batch dimension and $i$ indexes the token position. $g_{\theta}$ is a position-aware multilayer perception (MLP), and $\sigma$ denotes the sigmoid function. The resulting probabilities form a \emph{token importance profile} over positions. Each token is stochastically retained via independent Bernoulli sampling 
$m_{j,i} \sim \mathrm{Bernoulli}(p_{j,i})$, and reconstruction is performed from the masked discrete latent sequence $z_{q,j} \odot m_j$, where $m_j = [m_{j,1}, m_{j-2}, \ldots, m_{j,L}]$ is the dropping mask for the image $x_j$. Since Bernoulli sampling is non-differentiable, we adopt a straight-through estimator (STE) to allow gradient flow during training.

We can also compute the expected number of retained tokens for image $x_j$ by:
\begin{equation}
    T_j = \sum_{i=0}^{L-1} p_{j,i},
\end{equation}
so token allocation becomes a differentiable function of the 
continuous importance landscape rather than a discrete selection problem, hence the name `soft tail-dropping'.

\noindent\textbf{Content-adaptive prior.}
To allocate tokens based on visual complexity, we impose
a prior that encourages the expected token count to increase
with perceptual difficulty. For each image $x_j$, we compute
its perceptual reconstruction error $L_{\text{perc},j}$ (e.g., LPIPS)
as a proxy for visual complexity, and correlate this quantity
with the expected token count $T_j$.
Across all samples in a batch, we optimize
\begin{equation}
\mathcal{L}_{\text{content}} = \bigl(1 - \mathrm{corr}(L_{\text{perc}}, T)\bigr)^2,
\end{equation}
which encourages a strong positive correlation between the
complexity proxy $L_{\text{perc}}$ and the allocated token
count $T$.
$\mathcal{L}_{\text{content}}$ acts as a \emph{content-adaptive prior}
on token counts, nudging the tokenizer to assign more tokens
to harder images and fewer tokens to simpler ones. 
By contrasting all samples in the batch, this loss encourages the model to properly allocate tokens without the need to normalize the complexity proxy precisely. In practice, we compute the correlation using the Pearson correlation coefficient.

\noindent \textbf{Decreasing Importance Prior.} To match the causal structure prior required for autoregressive generation, we regularize the importance profile toward a decreasing form by penalizing upward steps:
\begin{equation}
    \mathcal{L}_{\text{decrease}}
= \sum_{i=1}^{L-1} \max\big(0,\, p_{j,i} - p_{j,i-1}\big).
\end{equation}
This encourages a soft monotonic decay along positions, stabilizes variable-length decoding, and produces consistent, content-dependent early-stop behavior when the tokenizer is used in AR models for generation.

\begin{figure}[!t]
\centering
\includegraphics[width=0.5\textwidth]{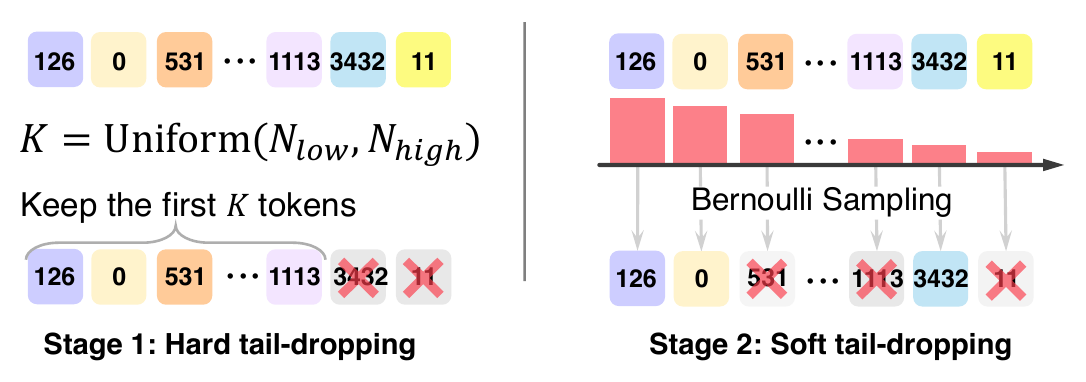}
\caption{Token dropping strategies in two training stages.}
\label{fig:dropping}
\end{figure}

\noindent \textbf{KL Sparsity Prior.}
The content adaptive loss shapes \emph{relative} allocation but does not enforce a global token budget. We impose a Bernoulli sparsity prior by regularizing the average keep probability:
\begin{equation}
\bar{p}_j = \frac{1}{L}\sum_{i=0}^{L-1} p_{j,i}
\end{equation}
toward a target preset sparsity $p^*$ using KL divergence:
\begin{equation}
\mathcal{L}_{\text{sparse}}
= \mathrm{KL}\!\left(\mathrm{Bern}(p^{*}) ~\|~ \mathrm{Bern}(\bar{p}_j)\right).
\end{equation}


\noindent \textbf{Overall Objective}
The adaptive tokenizer is trained with the composite loss
\begin{equation}
\begin{split}
\mathcal{L}
= \mathcal{L}_{\text{recon}} + \mathcal{L}_{\text{GAN}} + \mathcal{L}_{\text{VQ}}
+ \lambda_{\text{content}} \mathcal{L}_{\text{content}}  \\
+ \lambda_{\text{decrease}} \mathcal{L}_{\text{decrease}}
+ \lambda_{\text{sparse}} \mathcal{L}_{\text{sparse}}.
\end{split}
\end{equation}
This yields a tokenizer that is content-adaptive, contrastively aligned with visual complexity, structurally monotonic for AR usage, and globally sparse under a controllable token budget. Further details on all loss terms and the corresponding weight settings are provided in the supplementary material.

\subsection{Visual Generation with Adaptive Tokenizer}
The primary purpose of visual tokenizers is to provide an effective representation space for generative models. To evaluate this capacity, we integrate \ours{} with a vanilla autoregressive (AR) model and train the AR model on the discrete latent tokens produced by \ours{}. 

\noindent\textbf{1-D autoregressive modeling.}
Given an image $x$, \ours{} produces a quantized latent sequence
$q = (q_0, \ldots, q_{L-1})$. The AR model is then trained to model the distribution over this sequence as:
\begin{equation}
p_\phi(z_q) = \prod_{t=0}^{L} p_\phi(z_t \mid z_{<t}),
\end{equation}
where $p_\phi(\cdot)$ is the probability distribution parameterized by the AR model, and $\phi$ represents its learnable parameters.

\noindent\textbf{EoS from keep probabilities.}
Different from traditional fixed-length tokenizers, \ours{} encodes each image together with a keep-probability profile $p = (p_0, \ldots, p_{L-1})$, which is supposed to be nearly monotonic via $\mathcal{L}_{\text{decrease}}$ during the training of \ours{} (Sec.~\ref{sec:adaptive_tokens}). This structure allows the AR model to operate in an adaptive manner using a special end-of-sequence (EoS) token. During training, we randomly sample a probability threshold $\tau$ and define the EoS position as:
\begin{equation}
k = \min\{ i \mid p_i < \tau \}.
\end{equation}
The AR model is then supervised on the prefix $(z_0, \ldots, z_{k-1})$ followed by the EoS token. The monotonicity of $p$ ensures that the thresholding is stable and naturally aligned with the soft tail-dropping behavior of \ours{}.

\section{Experiments}
In this section, we evaluate the reconstruction capacity of \ours{}, and explore its integration with an autoregressive (AR) generation model, assessing its effectiveness in both class-conditional and text-conditional image generation tasks. We show that \ours{} achieves higher reconstruction quality while using fewer tokens than existing visual tokenizers, demonstrating its ability to perform content-aware compression. For image generation, we demonstrate that a vanilla AR model combined with \ours{} attains competitive performance against both diffusion-based approaches and AR models that rely on complex training pipelines. 
Finally, we extend our study to the video domain and validate the effectiveness of the soft probabilistic tail-dropping mechanism for adaptive video tokenization.

\noindent\textbf{Training setup. } We train \ours{} on the ImageNet-1k~\cite{deng2009imagenet} training set at a resolution of $256 \times 256$ using random-crop augmentation. 
The patch-embedding downsampling factor is set to $f = 16$, and the vector-quantizer uses a codebook of size 4096 with a code dimension of 12. The latent sequence length is set to 256. In the first stage, the keep length is sampled from a uniform distribution with bounds $L_{\text{min}} = 160$ and $L_{\text{max}} = 256$. At inference time, we use a default probability threshold of 0.5, retaining all tokens whose predicted keep probabilities exceed this threshold and dropping the rest. For additional hyperparameter settings and details of AR model training and inference, please refer to the supplementary material.

\noindent\textbf{Metrics.} We report reconstruction FID (rFID)~\cite{heusel2017gans} and PSNR for the image reconstruction task. For image generation, we evaluate generation FID (gFID)~\cite{heusel2017gans}, Inception Score (IS)~\cite{salimans2016improved}, and Precision/Recall~\cite{kynkaanniemi2019improved}. For video reconstruction, we measure reconstruction FVD~\cite{unterthiner2018towards} (rFVD), PSNR, and LPIPS~\cite{zhang2018unreasonable} to assess the quality.

\subsection{Image Reconstruction}
\begin{figure*}[!t]
\centering
\includegraphics[width=1.0\textwidth]{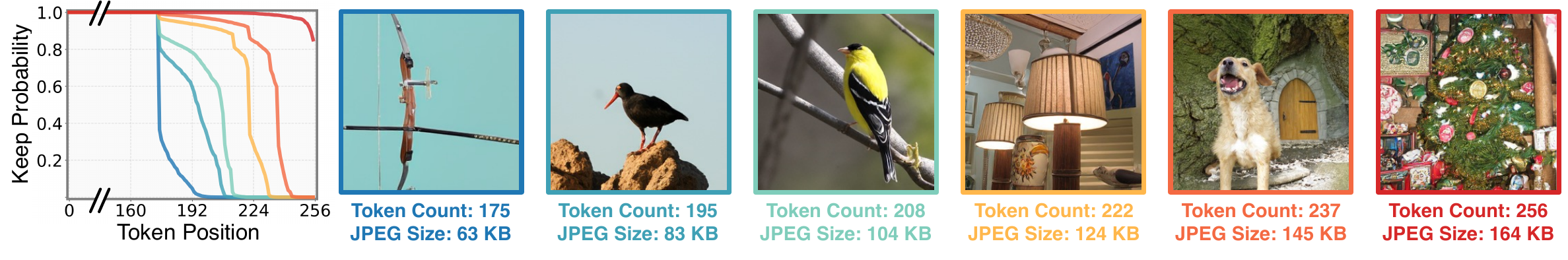}
\caption{Keep-probability curves and corresponding reconstructions across samples of varying complexity. More complex images receive higher token counts, and the learned allocation correlates strongly with the JPEG file sizes of the inputs.}
\label{fig:qualitative}
\end{figure*}

The evaluation for image reconstruction is conducted on the ImageNet-1K validation set at a resolution of $256\times256$. We compare \ours{} with representative 2D and 1D tokenizers. As shown in \cref{tables/img_recon}, \ours{} achieves the best rFID among all tokenizers while using only 90$\%$ of the token count compared to other tokenizers. This highlights the benefits from the adaptive token allocation strategy learned by \ours{}, which maximizes the representational efficiency of the available tokens. Moreover, when allowing more token budgets for reconstruction by either adjusting the probability threshold in inference or manually adding the budgets, \ours{} shows further improvements in rFID, albeit with a marginal decrease in PSNR. We hypothesize that this trade-off emerges because more tokens encourage the model to prioritize perceptual realism over pixel-level fidelity.

In \cref{fig:qualitative}, we show the keep-probability profile for different images. We show that \ours{} successfully learns a complexity-aware token allocation strategy, which is further validated by the high correlation between the allocated token counts and the JPEG sizes for storing the images.

\begin{table}[h]
\caption{\textbf{Comparison on ImageNet-1k image reconstruction.} We compare with 2D image tokenizers including Taming VQ-GAN~\cite{esser2021taming}, MaskGIT VQ-GAN~\cite{chang2022maskgit}, Open MAGVIT-v2~\cite{luo2024open}, and the tokenizer in LlamaGen~\cite{sun2024autoregressive}; 1D image tokenizers including TiTok~\cite{yu2024image}, FlexTok~\cite{bachmann2025flextok}, and One-D-Piece~\cite{miwa2025one}.
All evaluations are performed on the ImageNet-1k validation set at a resolution of 256$\times$256. \ours{} achieves the best rFID while using 10$\%$ fewer tokens compared with other tokenizers.  We additionally report results using a probability threshold of 0.01 at inference.
“(+X Tokens)” denotes manually adding X extra tokens to all images at inference. These variants of \ours{} further improves to an rFID of 0.88 when using an average of 240 tokens.}
\label{tables/img_recon}
\vspace{-0.5em}
\centering
\resizebox{\columnwidth}{!}{%
\begin{tabular}{c lcccc@{}}
\toprule
\textbf{Type} & \textbf{Tokenizer} & \textbf{\#Tokens} & \textbf{Codebook Size} & \textbf{rFID} $\downarrow$ & \textbf{PSNR} $\uparrow$ \\ 
\midrule
2D Tokenizer & Taming VQ-GAN~\cite{esser2021taming} & 256 & 16384 & 4.98 & 19.40 \\
& MaskGIT VQ-GAN~\cite{chang2022maskgit} & 256 & 1024 & 2.28 & - \\
& Open MAGVIT-v2~\cite{luo2024open} & 256 & 262144 & 1.17 & 22.64 \\
& LlamaGen~\cite{sun2024autoregressive} & 256 & 4096 & 3.02 & 19.99  \\ 
& LlamaGen~\cite{sun2024autoregressive} & 256 & 16384 & 2.19 & 20.79  \\ 
\midrule
1D Tokenizer & TiTok-L~\cite{yu2024image} & 32 & 4096 & 2.21 & 15.96  \\
& TiTok-B~\cite{yu2024image} & 64 & 4096 & 1.70 & 17.13 \\
& TiTok-S~\cite{yu2024image} & 128 & 4096 & 1.71 & 17.80 \\
& FlexTok~\cite{bachmann2025flextok} & 256 & 64000 & 1.08 & 17.70 \\
& One-D-Piece-L~\cite{miwa2025one} & 256 & 4096 & 1.08 & 19.04 \\
\midrule
1D Tokenizer & \ours-Fixcount & 220 & 4096 & 1.15 & 20.35 \\
& \ours-Harddrop & 222 & 4096 & 1.15 & 20.30 \\
\midrule
1D Tokenizer & \ours & 220 & 4096 & 1.15 & 20.22  \\
& \ours (Threshold 0.01) & 230 & 4096 & 0.99 & 20.25 \\
& \ours (Threshold 0.01; +10 Tokens) & 240 & 4096 & 0.88 & 20.02  \\ 
\bottomrule
\end{tabular}
}
\end{table} 

\subsection{Image Generation}
Given the strong reconstruction capability of \ours{}, we investigate whether this translates into improved generation performance. To this end, we integrate \ours{} into a vanilla autoregressive image generation framework, using the same backbone as LlamaGen~\cite{sun2024autoregressive}. An End-of-Sequence (EoS) token is used to allow the AR model to dynamically determine the effective token length during inference. 

As summarized in \cref{tables/img_gen}, the vanilla AR model equipped with \ours{} achieves generation quality that is comparable to state-of-the-art AR and diffusion-based models, despite its minimalist design. This demonstrates that \ours{} is well suited for AR generative modeling, improving both efficiency and flexibility without requiring elaborate architectural heuristics. These results suggest that \ours{} not only excels at reconstruction, but also enhances generation in a simple AR setup, underscoring its potential as a strong tokenizer for generation and unified multimodal models.

In \cref{tables/img_gen_tokenizers}, we compare different visual tokenizers within the same AR generation framework, including the improved VQ tokenizer from LlamaGen, TiTok-S, and One-D-Piece-L. \ours{} delivers the best generation performance, surpassing all baseline tokenizers by a substantial margin. These results further confirm the effectiveness of \ours{} in enhancing autoregressive image generation.

\begin{table*}[h]
\centering
\caption{\textbf{Comparison on ImageNet-1k class-conditional image generation.} We compare the AR generative model built on \ours{} with a broad range of generative models, including: GAN-based approaches such as BigGAN~\cite{brock2018large}, GigaGAN~\cite{kang2023scaling}, and StyleGAN~\cite{sauer2022stylegan}; diffusion-based models including ADM~\cite{dhariwal2021diffusion}, LDM~\cite{rombach2022high}, DiT~\cite{peebles2023scalable}, and SiT~\cite{ma2024sit}; autoregressive models with bi-directional attention including MaskGIT~\cite{chang2022maskgit}, MAGVIT-v2~\cite{yu2023language}, MAR~\cite{li2024autoregressive}, and TiTok~\cite{yu2024image}; autoregressive models with causal attention including VQGAN~\cite{esser2021taming}, RQTransformer~\cite{lee2022autoregressive}, Open-MAGVIT-v2~\cite{luo2024open}, VAR~\cite{tian2024visual}, SAR~\cite{liu2024customize}, RAR~\cite{yu2024randomized}, RandAR~\cite{pang2025randar}, and LlamaGen~\cite{sun2024autoregressive}. All evaluations are conducted on the ImageNet-1k validation set at a resolution of 256$\times$256. By simply replacing the original image tokenizer in LlamaGen with our proposed \ours{}, the 3B model achieves a gFID of 1.77, improving upon the LlamaGen baseline by 0.41 while using less than half the number of tokens. The performance of our AR model is also comparable to diffusion-based models and other AR models that rely on more advanced architectures and training strategies.} 
\label{tables/img_gen}

\resizebox{0.9\textwidth}{!}{
\begin{tabular}{c l c c c c c c}
\toprule
\textbf{Type} & \textbf{Model} & \textbf{\#Para.} & \textbf{gFID}$\downarrow$ & \textbf{IS}$\uparrow$ & \textbf{Precision}$\uparrow$ & \textbf{Recall}$\uparrow$  & \textbf{\#Tokens} \\
\midrule
GAN & BigGAN~\cite{brock2018large} & 112M & 6.95 & 224.5 & 0.89 & 0.38 & 1\\
& GigaGAN~\cite{kang2023scaling} & 569M & 3.45 & 225.5 & 0.84 & 0.61 & 1 \\
& StyleGAN-XL~\cite{sauer2022stylegan} & 166M & 2.30 & 265.1 & 0.78 & 0.53 & 1 \\
\midrule
Diffusion & ADM~\cite{dhariwal2021diffusion} & 554M & 4.59 & 186.7 & 0.82 & 0.53 & 250 \\
& LDM-4~\cite{rombach2022high} & 400M & 3.60 & 247.7 & -- & -- & 250\\
& DiT-XL~\cite{peebles2023scalable} & 675M & 2.27 & 278.2 & 0.83 & 0.57 & 250\\
& SiT-XL~\cite{ma2024sit} & 675M & 2.06 & 270.3 & 0.82 & 0.59 & 250 \\
\midrule
Bi-directional AR & MaskGIT-re \cite{chang2022maskgit} & 227M & 4.02 & 355.6 & -- & -- & 256 \\
& MAGVIT-v2~\cite{yu2023language} & 307M & 1.78 & 319.4 & -- & -- & 256 \\
& MAR-L~\cite{li2024autoregressive} & 479M & 1.98 & 290.3 & -- & -- & 64 \\
& MAR-H~\cite{li2024autoregressive} & 943M & 1.55 & 303.7 & 0.81 & 0.62 & 256 \\
& TiTok-S-128~\cite{yu2024image} & 287M & 1.97 & 281.8 & -- & -- & 128 \\
\midrule
Advanced Causal AR & VAR~\cite{tian2024visual} & 600M & 2.57 & 302.6 & 0.83 & 0.56 & 680 \\
& VAR~\cite{tian2024visual} & 2.0B & 1.92 & 350.2 & 0.82 & 0.59 & 680 \\
& SAR-XL~\cite{liu2024customize} & 893M & 2.76 & 273.8 & 0.84 & 0.55 & 256 \\
& RAR-XXL~\cite{yu2024randomized} & 955M & 1.50 & 306.9 & 0.80 & 0.62 & 256 \\
& RAR-XL~\cite{yu2024randomized} & 1.5B & 1.48 & 326.0 & 0.80 & 0.63 & 256 \\
& RandAR-XL~\cite{pang2025randar} & 775M & 2.25 & 317.8 & 0.80 & 0.60 & 256 \\ 
& RandAR-XXL~\cite{pang2025randar} & 1.4B & 2.15 & 322.0 & 0.79 & 0.62 & 256 \\
\midrule
Vanilla Causal AR & VQGAN~\cite{esser2021taming} & 1.4B & 5.20 & 280.3 & -- & -- & 256  \\
& RQTran.-re~\cite{lee2022autoregressive} & 3.8B & 3.80 & 323.7 & -- & -- & 256 \\
& Open-MAGVIT-v2-XL~\cite{luo2024open} & 1.5B & 2.33 & 271.8 & 0.84 & 0.54 & 256\\
& LlamaGen-XL-256~\cite{sun2024autoregressive} & 775M & 3.39 & 227.1 & 0.81 & 0.54 & 256 \\
& LlamaGen-XXL-256~\cite{sun2024autoregressive}  & 1.4B & 3.09 & 253.6 & 0.82 & 0.53 & 576 \\
& LlamaGen-3B-256~\cite{sun2024autoregressive}  & 3.1B & 3.06 & 279.7 & 0.84 & 0.54 & 576 \\
& LlamaGen-XL-576~\cite{sun2024autoregressive} & 775M & 2.62 & 244.1 & 0.80 & 0.57 & 576 \\
& LlamaGen-XXL-576~\cite{sun2024autoregressive}  & 1.4B & 2.34 & 253.9 & 0.80 & 0.59 & 576 \\
& LlamaGen-3B-576~\cite{sun2024autoregressive}  & 3.1B & 2.18 & 263.3 & 0.81 & 0.58 & 576 \\
\midrule
Vanilla Causal AR
& LlamaGen-\ours-XL & 775M & 2.36 & 244.0 & 0.78 & 0.62 & 223    \\ 
& LlamaGen-\ours-XXL & 1.4B & 1.91 & 290.2 & 0.79 & 0.63 & 227 \\ 
& LlamaGen-\ours-3B & 3.1B & 1.75 & 300.6 & 0.80 & 0.64 & 229 \\
\bottomrule
\end{tabular}
}
\end{table*}

\begin{figure}[!t]
\centering
\includegraphics[width=0.52\textwidth]{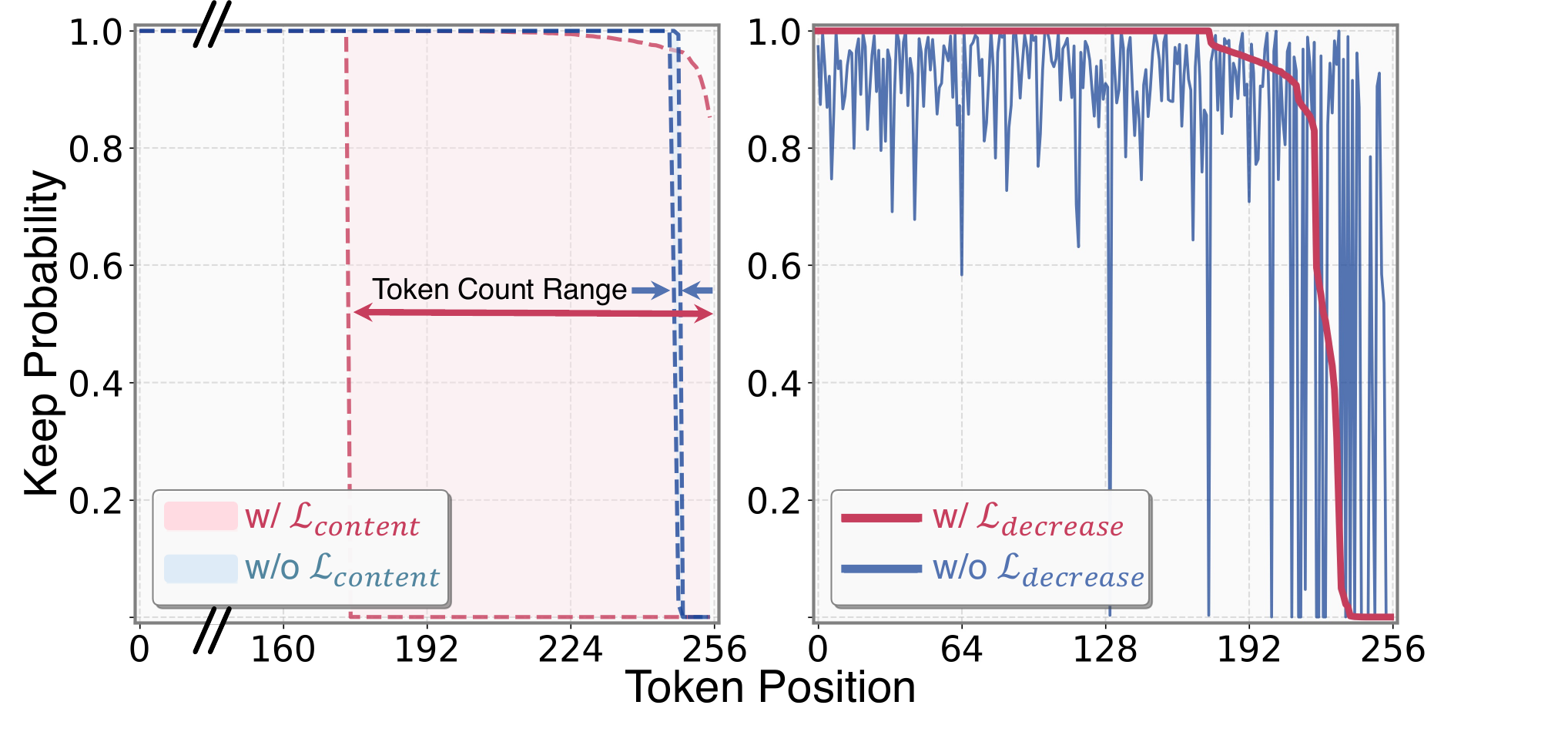}
\caption{Ablation Study on the two loss terms for decreasing importance prior and the content-adaptive prior. The decreasing importance prior is necessary for learning the monotonic structure for AR generation, while the content-adaptive prior is crucial for successfully learning token allocation strategy.}
\label{fig:ablation}
\end{figure}

\begin{table}[h]
\caption{\textbf{ImageNet-1k image class-conditional image generation performance on LlamaGen with different tokenizers.} We train LlamaGen, a vanilla transformer-based autoregressive model, on different tokenizers including the VQ tokenizer in~\cite{sun2024autoregressive}, TiTok~\cite{yu2024image}, One-D-Piece~\cite{miwa2025one}, and \ours{}. All experiments are conducted on LlamaGen-XL (775M) with the same training settings. The model trained with \ours{} outperforms other tokenizers by a large margin. We further evaluate a variant that replaces the original soft probabilistic dropping strategy with a deterministic tail-drop, and the results demonstrate that the soft probabilistic dropping mechanism is essential for achieving high-quality generation.}
\label{tables/img_gen_tokenizers}
\vspace{-0.5em}
\centering
\resizebox{\columnwidth}{!}{%
\begin{tabular}{lcccccc@{}}
\toprule
\textbf{Tokenizer} & \textbf{\#Tokens} & \textbf{Codebook Size} & \textbf{gFID} $\downarrow$ & \textbf{IS} $\uparrow$ & \textbf{Pre.} $\uparrow$ & \textbf{Rec.} $\uparrow$ \\ 
\midrule
LlamaGen~\cite{sun2024autoregressive} & 256 & 16384 & 3.39 & 227.1 & 0.81 & 0.54  \\
LlamaGen~\cite{sun2024autoregressive} & 576 & 16384 & 2.62 & 244.1 & 0.80 & 0.57  \\ 
TiTok-S~\cite{yu2024image} & 128 & 4096 & 4.90 & 191.7 & 0.77 & 0.56 \\
One-D-Piece-L~\cite{miwa2025one} & 256 & 4096 & 2.99 & 235.1 & 0.81 & 0.59 \\
\midrule
\ours-FixCount & 220 & 4096 & 2.73 & 234.2 & 0.77 & 0.62  \\
\ours-Harddrop & 223 & 4096 & 2.67 & 231.2 & 0.76 & 0.63  \\
\ours-Fixthreshold & 221 & 4096 & 2.49 & 240.7 & 0.79 & 0.61 \\
\midrule
\ours & 223 & 4096 & 2.36 & 244.0 & 0.78 & 0.62  \\
\bottomrule
\end{tabular}
}
\end{table}

\subsection{Ablation Study}
We investigate the effectiveness of different design components in the \ours{} pipeline, evaluating their impact on reconstruction and generation quality. 

\noindent\textbf{Adaptive Token Allocation.} To assess whether the learned adaptivity contributes to performance gains, we include an ablation in which the tokenizer uses a fixed token count for all images. This count is set to the average number of tokens used by \ours{} in inference when the probability threshold is 0.5. We denote this variant as \ours{}-Fixcount. 

The reconstruction comparison is shown in \cref{tables/img_recon}. While \ours{}-Fixcount achieves similar rFID and slightly higher PSNR, the adaptive version offers greater flexibility, enabling it to leverage additional token budgets to obtain improved rFID. For image generation results in \cref{tables/img_gen_tokenizers}), learning adaptivity leads to substantial performance gains. We attribute this improvement to that adaptive tokenization allows the AR model to jointly explore the latent code space and the token-count dimension, ultimately yielding higher-quality generations.

\noindent\textbf{Per-Token v.s. Per-Sample Dropping.} In \ours{}, token dropping is model as a per-token sampling process, and the proposed regularization terms shape the keep-probability profile into a soft tail-dropping pattern. To better understand its effect, we implement a hard token dropping variant, dubbed \ours{}-Harddrop, in which tokens are always truncated from the tail, and the token count is determined by the sum of the predicted keep probabilities across all token positions.

As shown in~\cref{tables/img_recon}, the two mechanisms show comparable performance, with \ours{}-Harddrop achieving slightly higher PSNR possibly due to the hard tail dropping pattern is easier for decoding. However, in image generation, \ours{} notably outperforms its harddrop counterpart. We attribute this to the stochasticity introduced by per-token sampling where early tokens are also possibly dropped. This stochastic perturbation improves robustness to imperfect tokens and effectively mitigates the exposure bias commonly observed in autoregressive inference.

\noindent\textbf{Position of the End-of-Sequence (EoS) Token. } For the AR generation model, we examine whether sampling the EoS threshold and allowing its position to vary across different passes of the same images could lead to better performance than using a fixed threshold that enforces a static EoS position for each image. As in \cref{tables/img_gen_tokenizers} between \ours{} and \ours{}-Fixthreshold, sampling the EoS threshold leads to a noticeable improvement in generation. The observation shows that the varying EoS positions could work as an augmentation for further improvement in generation.

\noindent\textbf{Regularization Losses.} We test the effectiveness of two loss terms: the decreasing loss and content-adaptive loss. As shown in~\cref{fig:ablation}, without the content-adaptive loss, the tokenizer is unable to learn effective token allocation strategies and learns only trivial solution where all images use almost the same token count. The decreasing loss is essential for the tokenizer to learn the soft tail dropping pattern and the structure prior of causality, which is necessary for AR generation training.

\subsection{Text-conditional Image Generation} We evaluate \ours{} on text-conditional image generation by comparing it with the enhanced VQ tokenizer from LlamaGen~\cite{sun2024autoregressive}.
Both tokenizers are trained on ImageNet, and we train separate vanilla autoregressive models with the two tokenizers on a subset of LAION-2B~\cite{schuhmann2022laion} for one epoch at the resolution 256$\times$256.
We measure the text-conditional generation quality on the GenEval~\cite{ghosh2023geneval} benchmark.

\begin{table}[h]
\caption{\textbf{Comparison between \ours{} and VQGAN from LlamaGen on Text-conditional generation results on GenEval.}}
\label{tables/img_gen_t2i}
\vspace{-0.5em}
\centering
\resizebox{\columnwidth}{!}{%
\begin{tabular}{l cccccc c@{}}
\toprule
\textbf{Tokenizer} & \textbf{S. Obj.} & \textbf{T. Obj.} & \textbf{Count.} & \textbf{Colors} & \textbf{Position} & \textbf{C. Attri.} & \textbf{Overall} \\ 
\midrule
LlamaGen & 0.91 & 0.18 & 0.19 & 0.57 & 0.04 & 0.01 & 0.32 \\
\ours & 0.99 & 0.36 & 0.40 & 0.81 & 0.10 & 0.06 & 0.45 \\
\bottomrule
\end{tabular}
}
\end{table}
\vspace{2.5mm}

\begin{table}[h]
\caption{\textbf{Comparison on UCF-101 video reconstruction.} We compare \ours-Video with other video tokenizers including TATS~\cite{ge2022long}, OmniTokenizer~\cite{wang2024omnitokenizer}, ElasticTok~\cite{yan2024elastictok}, Cosmos-Tokenizer-DV~\cite{agarwal2025cosmos}, LARP~\cite{wang2024larp}, and AdapTok~\cite{li2025learning}. Following the evaluation protocol of AdapTok~\cite{li2025learning}, all methods are evaluated on 16-frame video clips at a resolution of 128$\times$128. Our approach achieves the best PSNR and LPIPS while delivering competitive FVD with other state-of-the-art video tokenizers. ``(+ X Tokens)'' denotes manually adding X extra tokens to all videos at inference. $^\dagger$ indicates the baseline was trained for an additional 75 epochs.}
\label{tables/vid_recon}
\vspace{-1em}
\centering
\resizebox{\columnwidth}{!}
{%
\begin{tabular}{lccccc@{}}
\toprule
\textbf{Tokenizer} & \textbf{\#Tokens} & \textbf{Codebook Size} & \textbf{rFVD} $\downarrow$ & \textbf{PSNR} $\uparrow$ & \textbf{LPIPS} $\downarrow$ \\ 
\midrule
TATS~\cite{ge2022long} & 1024 & 16384 & 157 & 24.22 & 0.206 \\
OmniTokenizer~\cite{wang2024omnitokenizer} & 1280 & 8192 & 62 & 27.76 & 0.112 \\
ElasticTok~\cite{yan2024elastictok} & 1022 & 64000 & 230 & - & - \\
Cosmos-Tokenizer-DV~\cite{agarwal2025cosmos} & 1280 & 64000 & 140 & - & - \\
LARP~\cite{wang2024larp} & 1024 & 8192 & 24 & - & - \\
LARP$^\dagger$ ~\cite{wang2024larp} & 1024 & 8192 & 20 & 28.71 & 0.075\\
AdapTok~\cite{li2025learning} & 1024 & 8192 & 36 & 25.72 & 0.114 \\
\midrule
\ours-Video & 997 & 8192 & 29 & 28.77 & 0.064 \\
\ours-Video (+100 Tokens) & 1097 & 8192 & 26 & 29.08 & 0.062 \\
\ours-Video (+200 Tokens) & 1197 & 8192 & 23 & 29.27 & 0.060 \\
\ours-Video (+300 Tokens) & 1297 & 8192 & 22 & 29.40 & 0.059 \\
\bottomrule
\end{tabular}
}
\end{table}

As shown in \cref{tables/img_gen_t2i}, \ours{} consistently outperforms the VQ model across all metrics, demonstrating stronger generalization for text-conditional generation.

\subsection{Video Reconstruction}
We extend our soft tail-dropping strategy to the video domain by applying \ours{} to 16-frame video clips. As shown in \cref{tables/vid_recon}, \ours{}-Video achieves state-of-the-art reconstruction quality, obtaining the best PSNR and LPIPS among all compared video tokenizers, while also delivering competitive rFVD to prior state-of-the-art methods. Moreover, \ours{}-Video highlights the flexibility of adaptive tokenization: increasing the token budget consistently improves performance across all metrics, indicating that the proposed adaptive mechanism transfers effectively to the video domain and remains capable of learning content-dependent token allocation in spatiotemporal settings.

\section{Conclusion}
We present~\ours{}, a content-adaptive 1D visual tokenizer with probabilistic tail dropping that aligns token length with content complexity via a contrastive objective. This design allows~\ours{} to match or surpass state-of-the-art tokenizers in both reconstruction and generation while using fewer tokens. Our results show that adaptive, complexity-aware token cardinality serves as a strong inductive bias for high-dimensional generative modeling and could help enable unified multimodal autoregressive systems.
{
    \small
    \bibliographystyle{ieeenat_fullname}
    \bibliography{main}

@String(CVPR= {IEEE Conf. Comput. Vis. Pattern Recog.})

@String(ICCV= {Int. Conf. Comput. Vis.})

@String(ECCV= {Eur. Conf. Comput. Vis.})

@String(ICLR = {Int. Conf. Learn. Represent.})

@String(CVPR  = {CVPR})

@String(ICCV  = {ICCV})

@String(ECCV  = {ECCV})

@String(ICLR  = {ICLR})

@inproceedings{rombach2022high,
  title={High-resolution image synthesis with latent diffusion models},
  author={Rombach, Robin and Blattmann, Andreas and Lorenz, Dominik and Esser, Patrick and Ommer, Bj{\"o}rn},
  booktitle=CVPR,
  year={2022}
}

@inproceedings{peebles2023scalable,
  title={Scalable diffusion models with transformers},
  author={Peebles, William and Xie, Saining},
  booktitle=ICCV,
  year={2023}
}

@article{brock2018large,
  title={Large Scale GAN Training for High Fidelity Natural Image Synthesis},
  author={Brock, Andrew},
  journal={arXiv preprint arXiv:1809.11096},
  year={2018}
}

@inproceedings{kang2023scaling,
  title={Scaling up gans for text-to-image synthesis},
  author={Kang, Minguk and Zhu, Jun-Yan and Zhang, Richard and Park, Jaesik and Shechtman, Eli and Paris, Sylvain and Park, Taesung},
  booktitle={CVPR},
  year={2023}
}

@inproceedings{sauer2022stylegan,
  title={Stylegan-xl: Scaling stylegan to large diverse datasets},
  author={Sauer, Axel and Schwarz, Katja and Geiger, Andreas},
  booktitle={SIGGRAPH},
  year={2022}
}

@inproceedings{dhariwal2021diffusion,
  title={Diffusion models beat gans on image synthesis},
  author={Dhariwal, Prafulla and Nichol, Alexander},
  booktitle={NeurIPS},
  year={2021}
}

@inproceedings{ma2024sit,
  title={{SIT}: Exploring flow and diffusion-based generative models with scalable interpolant transformers},
  author={Ma, Nanye and Goldstein, Mark and Albergo, Michael S and Boffi, Nicholas M and Vanden-Eijnden, Eric and Xie, Saining},
  booktitle={ECCV},
  year={2024},
}

@inproceedings{chang2022maskgit,
  title={Maskgit: Masked generative image transformer},
  author={Chang, Huiwen and Zhang, Han and Jiang, Lu and Liu, Ce and Freeman, William T},
  booktitle={CVPR},
  year={2022}
}

@inproceedings{li2024autoregressive,
  title={Autoregressive Image Generation without Vector Quantization},
  author={Li, Tianhong and Tian, Yonglong and Li, He and Deng, Mingyang and He, Kaiming},
  booktitle={NeurIPS},
  year={2024}
}

@article{yu2024image,
  title={An image is worth 32 tokens for reconstruction and generation},
  author={Yu, Qihang and Weber, Mark and Deng, Xueqing and Shen, Xiaohui and Cremers, Daniel and Chen, Liang-Chieh},
  journal={NeurIPS},
  year={2024}
}

@inproceedings{esser2021taming,
  title={Taming transformers for high-resolution image synthesis},
  author={Esser, Patrick and Rombach, Robin and Ommer, Bjorn},
  booktitle={CVPR},
  year={2021}
}

@inproceedings{lee2022autoregressive,
  title={Autoregressive image generation using residual quantization},
  author={Lee, Doyup and Kim, Chiheon and Kim, Saehoon and Cho, Minsu and Han, Wook-Shin},
  booktitle={CVPR},
  year={2022}
}

@inproceedings{tian2024visual,
  title={Visual autoregressive modeling: Scalable image generation via next-scale prediction},
  author={Tian, Keyu and Jiang, Yi and Yuan, Zehuan and Peng, Bingyue and Wang, Liwei},
  booktitle={NeurIPS},
  year={2024}
}

@article{liu2024customize,
  title={Customize your visual autoregressive recipe with set autoregressive modeling},
  author={Liu, Wenze and Zhuo, Le and Xin, Yi and Xia, Sheng and Gao, Peng and Yue, Xiangyu},
  journal={arXiv preprint arXiv:2410.10511},
  year={2024}
}

@article{yu2024randomized,
  title={Randomized Autoregressive Visual Generation},
  author={Yu, Qihang and He, Ju and Deng, Xueqing and Shen, Xiaohui and Chen, Liang-Chieh},
  journal={arXiv preprint arXiv:2411.00776},
  year={2024}
}

@article{sun2024autoregressive,
  title={Autoregressive Model Beats Diffusion: Llama for Scalable Image Generation},
  author={Sun, Peize and Jiang, Yi and Chen, Shoufa and Zhang, Shilong and Peng, Bingyue and Luo, Ping and Yuan, Zehuan},
  journal={arXiv preprint arXiv:2406.06525},
  year={2024}
}

@inproceedings{pang2025randar,
  title={Randar: Decoder-only autoregressive visual generation in random orders},
  author={Pang, Ziqi and Zhang, Tianyuan and Luan, Fujun and Man, Yunze and Tan, Hao and Zhang, Kai and Freeman, William T and Wang, Yu-Xiong},
  booktitle={CVPR},
  year={2025}
}

@article{luo2024open,
  title={Open-magvit2: An open-source project toward democratizing auto-regressive visual generation},
  author={Luo, Zhuoyan and Shi, Fengyuan and Ge, Yixiao and Yang, Yujiu and Wang, Limin and Shan, Ying},
  journal={arXiv preprint arXiv:2409.04410},
  year={2024}
}

@inproceedings{yu2023language,
  title={Language Model Beats Diffusion--Tokenizer is Key to Visual Generation},
  author={Yu, Lijun and Lezama, José and Gundavarapu, Nitesh B. and Versari, Luca and Sohn, Kihyuk and Minnen, David and Cheng, Yong and Birodkar, Vighnesh and Gupta, Agrim and Gu, Xiuye and Hauptmann, Alexander G. and Gong, Boqing and Yang, Ming-Hsuan and Essa, Irfan and Ross, David A. and Jiang, Lu},
  booktitle={ICLR},
  year={2024}
}

@article{miwa2025one,
  title={One-d-piece: Image tokenizer meets quality-controllable compression},
  author={Miwa, Keita and Sasaki, Kento and Arai, Hidehisa and Takahashi, Tsubasa and Yamaguchi, Yu},
  journal={arXiv preprint arXiv:2501.10064},
  year={2025}
}

@inproceedings{bachmann2025flextok,
  title={FlexTok: Resampling Images into 1D Token Sequences of Flexible Length},
  author={Bachmann, Roman and Allardice, Jesse and Mizrahi, David and Fini, Enrico and Kar, O{\u{g}}uzhan Fatih and Amirloo, Elmira and El-Nouby, Alaaeldin and Zamir, Amir and Dehghan, Afshin},
  booktitle={ICML},
  year={2025}
}

@article{li2025learning,
  title={Learning Adaptive and Temporally Causal Video Tokenization in a 1D Latent Space},
  author={Li, Yan and Tian, Changyao and Xia, Renqiu and Liao, Ning and Guo, Weiwei and Yan, Junchi and Li, Hongsheng and Dai, Jifeng and Li, Hao and Yang, Xue},
  journal={arXiv preprint arXiv:2505.17011},
  year={2025}
}

@article{yan2024elastictok,
  title={Elastictok: Adaptive tokenization for image and video},
  author={Yan, Wilson and Mnih, Volodymyr and Faust, Aleksandra and Zaharia, Matei and Abbeel, Pieter and Liu, Hao},
  journal={ICLR},
  year={2025}
}

@article{wang2024omnitokenizer,
  title={Omnitokenizer: A joint image-video tokenizer for visual generation},
  author={Wang, Junke and Jiang, Yi and Yuan, Zehuan and Peng, Bingyue and Wu, Zuxuan and Jiang, Yu-Gang},
  journal={NeurIPS},
  year={2024}
}

@article{agarwal2025cosmos,
  title={Cosmos world foundation model platform for physical ai},
  author={Agarwal, Niket and Ali, Arslan and Bala, Maciej and Balaji, Yogesh and Barker, Erik and Cai, Tiffany and Chattopadhyay, Prithvijit and Chen, Yongxin and Cui, Yin and Ding, Yifan and others},
  journal={arXiv preprint arXiv:2501.03575},
  year={2025}
}

@inproceedings{ge2022long,
  title={Long video generation with time-agnostic vqgan and time-sensitive transformer},
  author={Ge, Songwei and Hayes, Thomas and Yang, Harry and Yin, Xi and Pang, Guan and Jacobs, David and Huang, Jia-Bin and Parikh, Devi},
  booktitle={ECCV},
  year={2022},
}

@article{wang2024larp,
  title={Larp: Tokenizing videos with a learned autoregressive generative prior},
  author={Wang, Hanyu and Suri, Saksham and Ren, Yixuan and Chen, Hao and Shrivastava, Abhinav},
  journal={arXiv preprint arXiv:2410.21264},
  year={2024}
}

@inproceedings{deng2009imagenet,
  title={Imagenet: A large-scale hierarchical image database},
  author={Deng, Jia and Dong, Wei and Socher, Richard and Li, Li-Jia and Li, Kai and Fei-Fei, Li},
  booktitle={CVPR},
  year={2009},
}

@inproceedings{zhang2018unreasonable,
  title={The unreasonable effectiveness of deep features as a perceptual metric},
  author={Zhang, Richard and Isola, Phillip and Efros, Alexei A and Shechtman, Eli and Wang, Oliver},
  booktitle={CVPR},
  year={2018}
}

@article{touvron2023llama,
  title={Llama: Open and efficient foundation language models},
  author={Touvron, Hugo and Lavril, Thibaut and Izacard, Gautier and Martinet, Xavier and Lachaux, Marie-Anne and Lacroix, Timoth{\'e}e and Rozi{\`e}re, Baptiste and Goyal, Naman and Hambro, Eric and Azhar, Faisal and others},
  journal={arXiv preprint arXiv:2302.13971},
  year={2023}
}

@article{yu2021vector,
  title={Vector-quantized image modeling with improved vqgan},
  author={Yu, Jiahui and Li, Xin and Koh, Jing Yu and Zhang, Han and Pang, Ruoming and Qin, James and Ku, Alexander and Xu, Yuanzhong and Baldridge, Jason and Wu, Yonghui},
  journal={arXiv preprint arXiv:2110.04627},
  year={2021}
}

@inproceedings{mentzer2023finite,
  title={Finite scalar quantization: Vq-vae made simple},
  author={Mentzer, Fabian and Minnen, David and Agustsson, Eirikur and Tschannen, Michael},
  booktitle={ICLR},
  year={2024}
}

@inproceedings{kim2025democratizing,
  title={Democratizing text-to-image masked generative models with compact text-aware one-dimensional tokens},
  author={Kim, Dongwon and He, Ju and Yu, Qihang and Yang, Chenglin and Shen, Xiaohui and Kwak, Suha and Chen, Liang-Chieh},
  booktitle={ICCV},
  year={2025}
}

@inproceedings{duggal2024adaptive,
  title={Adaptive length image tokenization via recurrent allocation},
  author={Duggal, Shivam and Isola, Phillip and Torralba, Antonio and Freeman, William T},
  booktitle={ICLR},
  year={2025}
}

@inproceedings{duggal2025single,
  title={Single-pass adaptive image tokenization for minimum program search},
  author={Duggal, Shivam and Byun, Sanghyun and Freeman, William T and Torralba, Antonio and Isola, Phillip},
  booktitle={NeurIPS},
  year={2025}
}

@article{el2022image,
  title={Image compression with product quantized masked image modeling},
  author={El-Nouby, Alaaeldin and Muckley, Matthew J and Ullrich, Karen and Laptev, Ivan and Verbeek, Jakob and J{\'e}gou, Herv{\'e}},
  journal={TMLR},
  year={2023}
}

@inproceedings{ramesh2021zero,
  title={Zero-shot text-to-image generation},
  author={Ramesh, Aditya and Pavlov, Mikhail and Goh, Gabriel and Gray, Scott and Voss, Chelsea and Radford, Alec and Chen, Mark and Sutskever, Ilya},
  booktitle={ICML},
  year={2021},
}

@article{van2017neural,
  title={Neural discrete representation learning},
  author={Van Den Oord, Aaron and Vinyals, Oriol and others},
  journal={NeurIPS},
  year={2017}
}

@inproceedings{rippel2014learning,
  title={Learning ordered representations with nested dropout},
  author={Rippel, Oren and Gelbart, Michael and Adams, Ryan},
  booktitle={ICML},
  year={2014},
}

@article{goodfellow2014generative,
  title={Generative adversarial nets},
  author={Goodfellow, Ian J and Pouget-Abadie, Jean and Mirza, Mehdi and Xu, Bing and Warde-Farley, David and Ozair, Sherjil and Courville, Aaron and Bengio, Yoshua},
  journal={NeurIPS},
  year={2014}
}

@inproceedings{lezama2022improved,
  title={Improved masked image generation with token-critic},
  author={Lezama, Jos{\'e} and Chang, Huiwen and Jiang, Lu and Essa, Irfan},
  booktitle={ECCV},
  year={2022},
}

@inproceedings{ren2025beyond,
  title={Beyond next-token: Next-x prediction for autoregressive visual generation},
  author={Ren, Sucheng and Yu, Qihang and He, Ju and Shen, Xiaohui and Yuille, Alan and Chen, Liang-Chieh},
  booktitle={ICCV},
  year={2025}
}

@article{heusel2017gans,
  title={Gans trained by a two time-scale update rule converge to a local nash equilibrium},
  author={Heusel, Martin and Ramsauer, Hubert and Unterthiner, Thomas and Nessler, Bernhard and Hochreiter, Sepp},
  journal={NeurIPS},
  year={2017}
}

@article{unterthiner2018towards,
  title={Towards accurate generative models of video: A new metric \& challenges},
  author={Unterthiner, Thomas and Van Steenkiste, Sjoerd and Kurach, Karol and Marinier, Raphael and Michalski, Marcin and Gelly, Sylvain},
  journal={arXiv preprint arXiv:1812.01717},
  year={2018}
}

@article{salimans2016improved,
  title={Improved techniques for training gans},
  author={Salimans, Tim and Goodfellow, Ian and Zaremba, Wojciech and Cheung, Vicki and Radford, Alec and Chen, Xi},
  journal={NeurIPS},
  year={2016}
}

@article{kynkaanniemi2019improved,
  title={Improved precision and recall metric for assessing generative models},
  author={Kynk{\"a}{\"a}nniemi, Tuomas and Karras, Tero and Laine, Samuli and Lehtinen, Jaakko and Aila, Timo},
  journal={NeurIPS},
  year={2019}
}

@article{schuhmann2022laion,
  title={Laion-5b: An open large-scale dataset for training next generation image-text models},
  author={Schuhmann, Christoph and Beaumont, Romain and Vencu, Richard and Gordon, Cade and Wightman, Ross and Cherti, Mehdi and Coombes, Theo and Katta, Aarush and Mullis, Clayton and Wortsman, Mitchell and others},
  journal={NeurIPS},
  year={2022}
}

@article{ghosh2023geneval,
  title={Geneval: An object-focused framework for evaluating text-to-image alignment},
  author={Ghosh, Dhruba and Hajishirzi, Hannaneh and Schmidt, Ludwig},
  journal={NeurIPS},
  year={2023}
}

@article{kingma2013auto,
  title={Auto-encoding variational bayes},
  author={Kingma, Diederik P and Welling, Max},
  journal={arXiv preprint arXiv:1312.6114},
  year={2013}
}

@article{razavi2019generating,
  title={Generating diverse high-fidelity images with vq-vae-2},
  author={Razavi, Ali and Van den Oord, Aaron and Vinyals, Oriol},
  journal={NeurIPS},
  year={2019}
}

@inproceedings{sennrich2016BPE,
  title={Neural machine translation of rare words with subword units},
  author={Sennrich, Rico and Haddow, Barry and Birch, Alexandra},
  booktitle={ACL},
  year={2016}
}

@article{Radford2018GPT,
  title={Improving language understanding by generative pre-training},
  author={Radford, Alec and Narasimhan, Karthik and Salimans, Tim and Sutskever, Ilya and others},
  year={2018}
}

@inproceedings{chen20iGPT,
  title={Generative pretraining from pixels},
  author={Chen, Mark and Radford, Alec and Child, Rewon and Wu, Jeffrey and Jun, Heewoo and Luan, David and Sutskever, Ilya},
  booktitle={ICML},
  year={2020},
}

@misc{Tian2025Infinity,
    title={Infinity: Scaling Bitwise AutoRegressive Modeling for High-Resolution Image Synthesis}, 
    author={Jian Han and Jinlai Liu and Yi Jiang and Bin Yan and Yuqi Zhang and Zehuan Yuan and Bingyue Peng and Xiaobing Liu},
    year={2024},
    eprint={2412.04431},
    archivePrefix={arXiv},
    primaryClass={cs.CV},
    url={https://arxiv.org/abs/2412.04431}, 
}

@article{Zhou2024TransfusionPT,
  title={Transfusion: Predict the next token and diffuse images with one multi-modal model},
  author={Zhou, Chunting and Yu, Lili and Babu, Arun and Tirumala, Kushal and Yasunaga, Michihiro and Shamis, Leonid and Kahn, Jacob and Ma, Xuezhe and Zettlemoyer, Luke and Levy, Omer},
  journal={arXiv preprint arXiv:2408.11039},
  year={2024}
}

@article{deng2025bagel,
  title   = {Emerging Properties in Unified Multimodal Pretraining},
  author  = {Deng, Chaorui and Zhu, Deyao and Li, Kunchang and Gou, Chenhui and Li, Feng and Wang, Zeyu and Zhong, Shu and Yu, Weihao and Nie, Xiaonan and Song, Ziang and Shi, Guang and Fan, Haoqi},
  journal = {arXiv preprint arXiv:2505.14683},
  year    = {2025}
}

@article{ho2020denoising,
  title={Denoising Diffusion Probabilistic Models},
  author={Jonathan Ho and Ajay Jain and Pieter Abbeel},
  year={2020},
  journal={arXiv preprint arxiv:2006.11239}
}

@inproceedings{song2021ddim,
title={Denoising Diffusion Implicit Models},
author={Jiaming Song and Chenlin Meng and Stefano Ermon},
booktitle={ICLR},
year={2021},
}

@misc{cui2025emu35nativemultimodalmodels,
      title={Emu3.5: Native Multimodal Models are World Learners}, 
      author={Yufeng Cui and Honghao Chen and Haoge Deng and Xu Huang and Xinghang Li and Jirong Liu and Yang Liu and Zhuoyan Luo and Jinsheng Wang and Wenxuan Wang and Yueze Wang and Chengyuan Wang and Fan Zhang and Yingli Zhao and Ting Pan and Xianduo Li and Zecheng Hao and Wenxuan Ma and Zhuo Chen and Yulong Ao and Tiejun Huang and Zhongyuan Wang and Xinlong Wang},
      year={2025},
      eprint={2510.26583},
      archivePrefix={arXiv},
      url={https://arxiv.org/abs/2510.26583}, 
}

@article{radford2019language,
  title={Language Models are Unsupervised Multitask Learners},
  author={Radford, Alec and Wu, Jeff and Child, Rewon and Luan, David and Amodei, Dario and Sutskever, Ilya},
  year={2019}
}

@article{hendrycks2016gaussian,
  title={Gaussian Error Linear Units (Gelus)},
  author={Hendrycks, D},
  journal={arXiv preprint arXiv:1606.08415},
  year={2016}
}

@article{su2024roformer,
  title={Roformer: Enhanced transformer with rotary position embedding},
  author={Su, Jianlin and Ahmed, Murtadha and Lu, Yu and Pan, Shengfeng and Bo, Wen and Liu, Yunfeng},
  journal={Neurocomputing},
  year={2024},
}

@article{ho2022classifier,
  title={Classifier-free diffusion guidance},
  author={Ho, Jonathan and Salimans, Tim},
  journal={arXiv preprint arXiv:2207.12598},
  year={2022}
}

@inproceedings{gao2023masked,
  title={Masked diffusion transformer is a strong image synthesizer},
  author={Gao, Shanghua and Zhou, Pan and Cheng, Ming-Ming and Yan, Shuicheng},
  booktitle={ICCV},
  year={2023}
}

@inproceedings{kingma2014adam,
  title={Adam: A method for stochastic optimization},
  author={Kingma, Diederik P},
  booktitle={ICLR},
  year={2015}
}

@article{vaswani2017attention,
  title={Attention is all you need},
  author={Vaswani, Ashish and Shazeer, Noam and Parmar, Niki and Uszkoreit, Jakob and Jones, Llion and Gomez, Aidan N and Kaiser, {\L}ukasz and Polosukhin, Illia},
  journal={NeurIPS},
  year={2017}
}

@inproceedings{chen2025x,
  title={X-dancer: Expressive music to human dance video generation},
  author={Chen, Zeyuan and Xu, Hongyi and Song, Guoxian and Xie, You and Zhang, Chenxu and Chen, Xin and Wang, Chao and Chang, Di and Luo, Linjie},
  booktitle={ICCV},
  year={2025}
}

@inproceedings{mao2025sir,
  title={SIR-DIFF: Sparse Image Sets Restoration with Multi-View Diffusion Model},
  author={Mao, Yucheng and Wang, Boyang and Kulkarni, Nilesh and Park, Jeong Joon},
  booktitle={CVPR},
  year={2025}
}

@InProceedings{Chang_2025_CVPR,
    author    = {Chang, Di and Xu, Hongyi and Xie, You and Gao, Yipeng and Kuang, Zhengfei and Cai, Shengqu and Zhang, Chenxu and Song, Guoxian and Wang, Chao and Shi, Yichun and Chen, Zeyuan and Zhou, Shijie and Luo, Linjie and Wetzstein, Gordon and Soleymani, Mohammad},
    title     = {X-Dyna: Expressive Dynamic Human Image Animation},
    booktitle = {CVPR},
    year      = {2025},
}

@inproceedings{zeng2025yolo,
  title={Yolo-count: Differentiable object counting for text-to-image generation},
  author={Zeng, Guanning and Zhang, Xiang and Wang, Zirui and Xu, Haiyang and Chen, Zeyuan and Li, Bingnan and Tu, Zhuowen},
  booktitle={ICCV},
  year={2025}
}

@inproceedings{chen2025cvp,
  title={CVP: Central-Peripheral Vision-Inspired Multimodal Model for Spatial Reasoning},
  author={Chen, Zeyuan and Zhang, Xiang and Xu, Haiyang and Xie, Jianwen and Tu, Zhuowen},
  booktitle={WACV},
  year={2026}
}

@inproceedings{zhao2024bsqvit,
  title={Image and Video Tokenization with Binary Spherical Quantization},
  author={Zhao, Yue and Xiong, Yuanjun and Krahenbuhl, Philipp},
  booktitle={ICLR},
  year={2025}
}
}

\clearpage
\setcounter{page}{1}
\maketitlesupplementary

In the supplementary material, we begin with analyzing the token count distribution and its relationship to JPEG file size in \cref{sec:jpeg}. Next, we provide additional quantitative and qualitative results in Secs.~\ref{sec:ablation_inference}, \ref{sec:recon_512}, and \ref{sec:visualization}. Finally, we discuss the limitations of \ours{} in \cref{sec:limitation} and present full implementation details in \cref{sec:implementation}.

\section{Token Count and JPEG Size}
\label{sec:jpeg}
\begin{figure}[!h]
\centering
\includegraphics[width=0.52\textwidth]{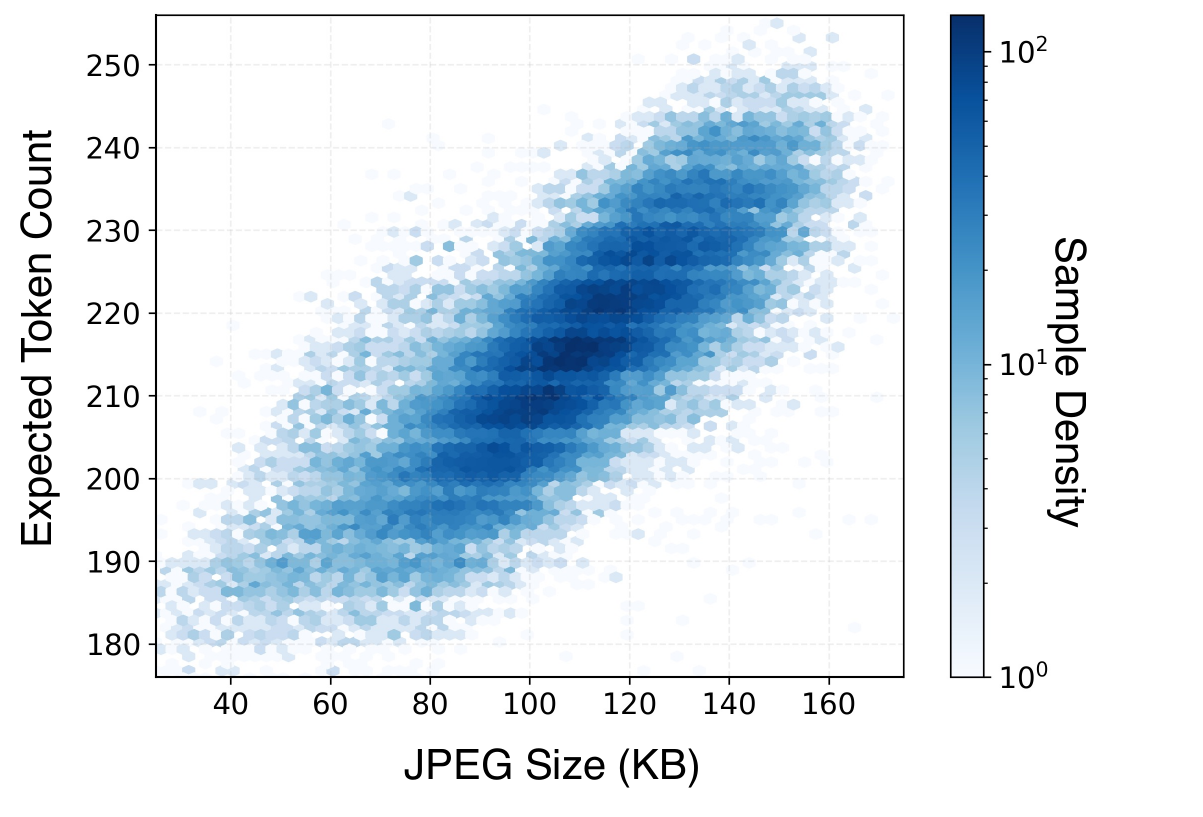}
\caption{Visualization of JPEG file size versus the expected token count predicted by our adaptive tokenizer. We evaluate 50,000 ImageNet-1k validation images, with each hexagonal bin showing the log-density of samples. The strong correlation indicates that \ours{} effectively aligns token allocation with the underlying image complexity reflected by JPEG size.}
\label{fig:jpeg_image}
\end{figure}

We conduct inference on the ImageNet-1k validation set, which contains 50,000 images. For each image, we record its JPEG file size and use \ours{} to compute the expected token count for decoding by summing the per-position keep probabilities. The visualization in \cref{fig:jpeg_image} shows a clear and strong correlation between JPEG size and the predicted token count. This implies that \ours{} implicitly models frequency-domain complexity—assigning more tokens to images with rich high-frequency content (e.g., textures, edges) and fewer tokens to smoother regions, despite not being trained with any explicit frequency supervision.

We further fit a linear regression to quantify this relationship and obtain the following equation between the expected token count and the JPEG file size:
\begin{equation}
    \text{Token-Count} = 0.44 \times \text{JPEG-Size (KB)} + 167.6.
\end{equation}
This suggests that increasing the token budget by one token corresponds to encoding roughly an additional 2 KB of information measured under the JPEG compression scheme.

\section{Ablation Study on \ours{} Inference}
\label{sec:ablation_inference}
We ablate how \ours{} determines the token count \emph{in inference} using the predicted per-token keep probabilities. By default, we apply a threshold of 0.5 and discard all tokens whose keep probabilities fall below this value. As an alternative, we compute the expected number of retained tokens by summing all keep probabilities for an image, and then keep only the first tokens up to this expected count, truncating the remaining tail tokens. Since \ours{} imposes a decreasing importance prior that enforces a monotonic decay in keep probabilities, the behaviors of the thresholding and expectation-based strategies are similar. The results are reported in \cref{tables/img_recon_ablation}, which show that once the average token counts are matched across the two inference strategies, their reconstruction performance is very close.

\begin{table}[h]
\caption{\textbf{Comparison on ImageNet-1k image reconstruction.} Ablation of inference-time settings in \ours{}.}
\label{tables/img_recon_ablation}
\vspace{-0.5em}
\centering
\resizebox{\columnwidth}{!}{%
\begin{tabular}{lccc@{}}
\toprule
\textbf{Tokenizer} & \textbf{\#Tokens} & \textbf{rFID} $\downarrow$ & \textbf{PSNR} $\uparrow$ \\ 
\midrule
\ours(threshold 0.5)& 220 & 1.15 & 20.22  \\
\ours(expected token count) & 216 & 1.20 & 20.21 \\
\ours(expected token count; +4 tokens) & 220 & 1.14 & 20.24 \\
\bottomrule
\end{tabular}
}
\end{table}

\begin{figure*}[!t]
\centering
\includegraphics[width=1.02\textwidth]{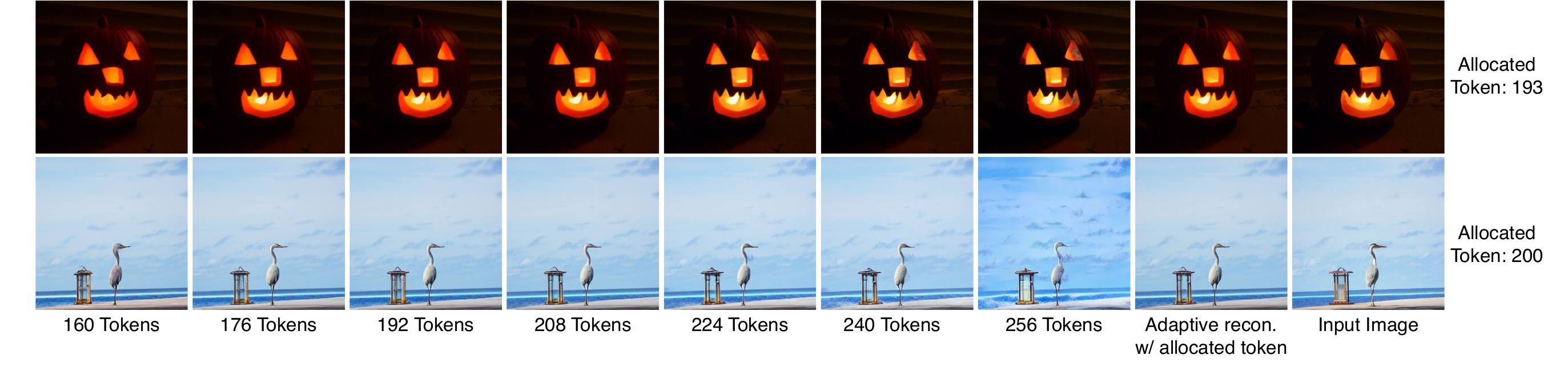}
\caption{\textbf{Limitation of \ours{}.} Increasing the token count beyond the amount allocated by \ours{} does not always improve reconstruction faithfulness, as seen in the artifacts in the pumpkin’s eye and the color distortions in the heron sample when reconstructed with 256 tokens.}
\label{fig:limitation}
\end{figure*}

\begin{table}[h]
\caption{\textbf{Comparison on ImageNet-1k image reconstruction at the resolution of 512$\times$512.}}
\label{tables/img_recon_512}
\vspace{-0.5em}
\centering
\resizebox{\columnwidth}{!}{%
\begin{tabular}{lcccc@{}}
\toprule
\textbf{Tokenizer} & \textbf{\#Tokens} & \textbf{Codebook Size} & \textbf{rFID} $\downarrow$ & \textbf{PSNR} $\uparrow$ \\ 
\midrule
MaskGIT VQ-GAN~\cite{chang2022maskgit} & 1024 & 1024 & 1.97 & - \\
TiTok-L-64~\cite{yu2024image} & 64 & 4096 & 1.77 & - \\
TiTok-B-128~\cite{yu2024image} & 128 & 4096 & 1.52 & - \\
LlamaGen~\cite{sun2024autoregressive} & 1024 & 16384 & 0.70 & 23.03 \\
\midrule
\ours(threshold 0.5)& 471 & 4096 & 0.87 & 20.98  \\
\ours(threshold 0.01) & 482 & 4096 & 0.82 & 20.98 \\
\ours(threshold 0.01; +20 tokens) & 501 & 4096 & 0.75 & 20.88 \\
\bottomrule
\end{tabular}
}
\end{table}
\section{Image Reconstruction at 512$\times$512}
\label{sec:recon_512}
We further evaluate the feasibility of \ours{} at a resolution of 512$\times$512 on ImageNet. All settings remain unchanged except that we double the latent sequence length to accommodate higher-resolution reconstructions. The results are reported in \cref{tables/img_recon_512}. \ours{} achieves an rFID competitive with the tokenizer used in LlamaGen~\cite{sun2024autoregressive}, while using a smaller codebook and less than half the average number of tokens. We note that the PSNR of \ours{} is lower than that of LlamaGen, likely because the model is bottlenecked by both the codebook size and the number of tokens, which limits its ability to recover all fine-grained details. Nevertheless, the low rFID indicates that \ours{} is still capable of producing high-quality reconstructions at a higher resolution.

\section{Qualitative Results}
\label{sec:visualization}
We present qualitative results on image reconstruction, class-conditional image generation, and text-conditional image generation from page \pageref{fig:recon} to page \pageref{fig:adaptive_gen}.

\subsection{Image Reconstruction}
\begin{figure*}[!t]
\centering
\includegraphics[width=0.97\textwidth]{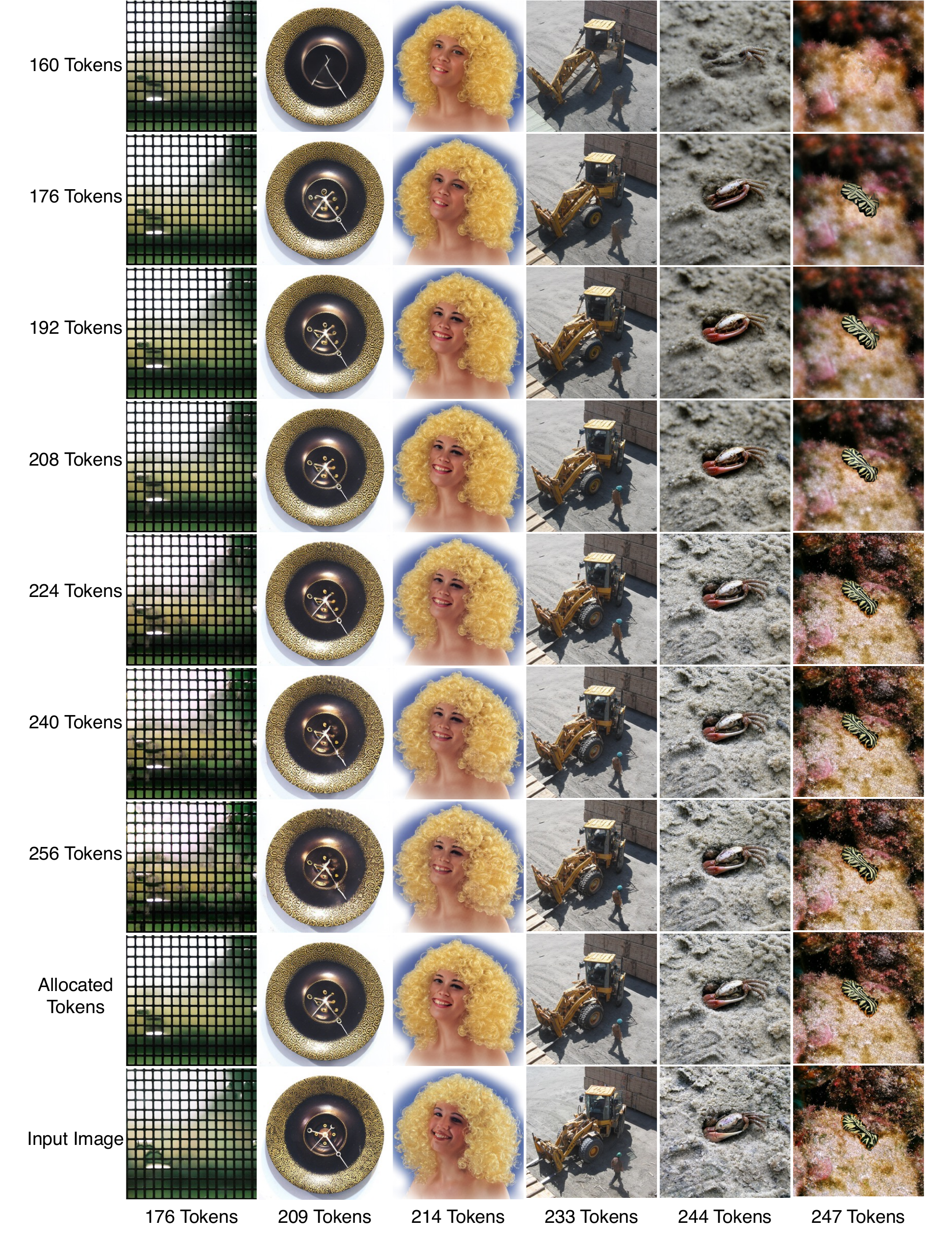}
\vspace{-2.5mm}
\caption{\textbf{Adaptive image reconstruction with \ours{}.} Bottom labels indicate the allocated tokens. Each row shows reconstructions using a fixed token count (left), while the last two rows depict results using the token counts predicted by \ours{} and the original input images.}
\label{fig:recon}
\end{figure*}

The reconstruction results of \ours{} with varying token counts are shown in \cref{fig:recon}. As illustrated, \ours{} naturally assigns larger token counts to visually complex images. From top to bottom, reconstructions with fewer tokens exhibit blurry regions or missing fine details (e.g., the rightmost two samples), while increasing the token counts progressively restores clear structures and sharp textures.

\subsection{Text-conditional Image Generation}
\begin{figure*}[!t]
\centering
\includegraphics[width=1.0\textwidth]{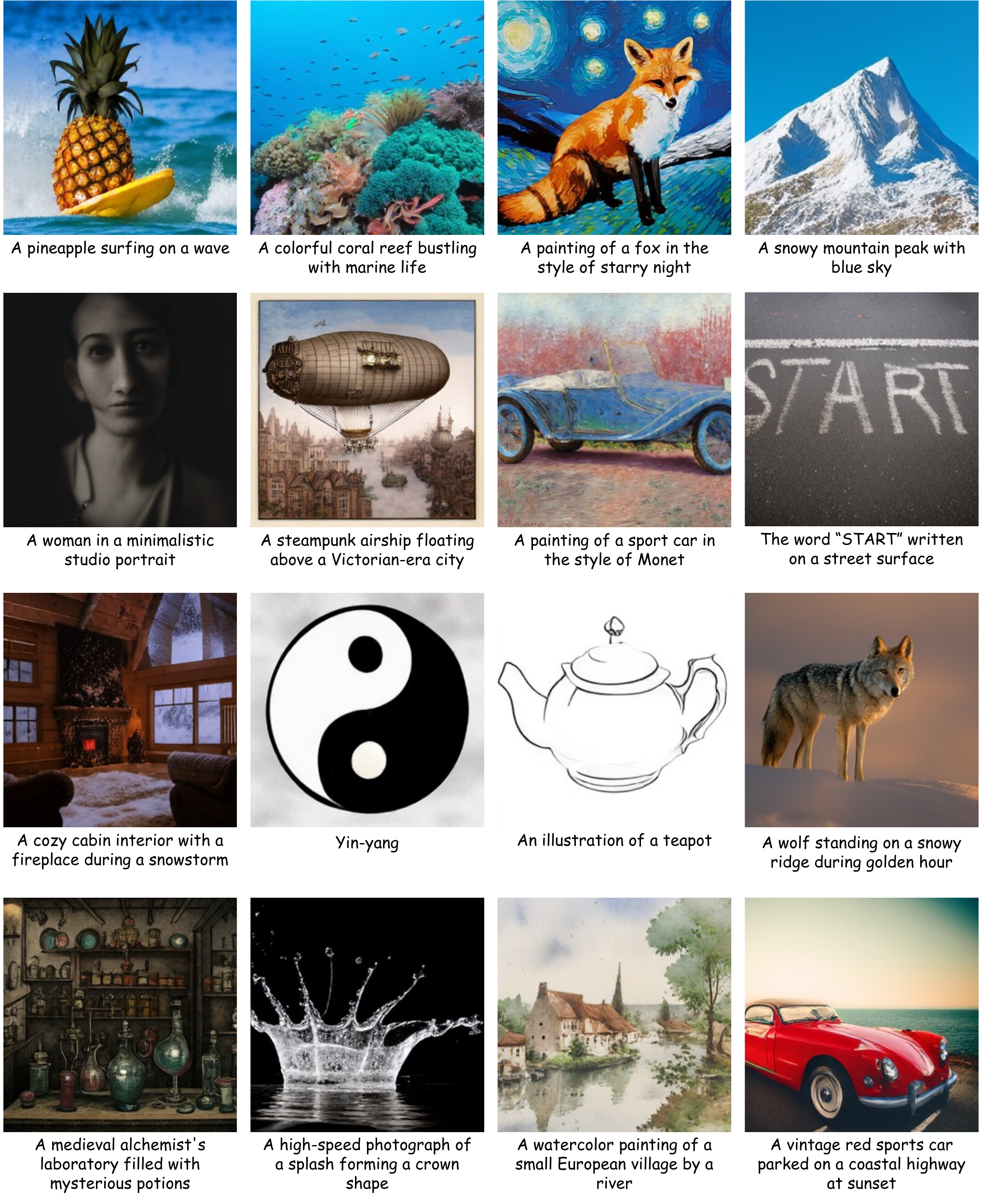}
\caption{\textbf{Text-conditional 256$\times$256 Image Generation.} The input text prompts are shown below the images.}
\label{fig:t2i}
\end{figure*}

We show text-conditional image generation results in \cref{fig:t2i}. All generations are at a resolution of 256$\times$256. The results demonstrate that a vanilla autoregressive model paired with \ours{} achieves strong text-conditional generation quality.

\subsection{Class-conditional Image Generation}
\begin{figure*}[!t]
\centering
\includegraphics[width=0.93\textwidth]{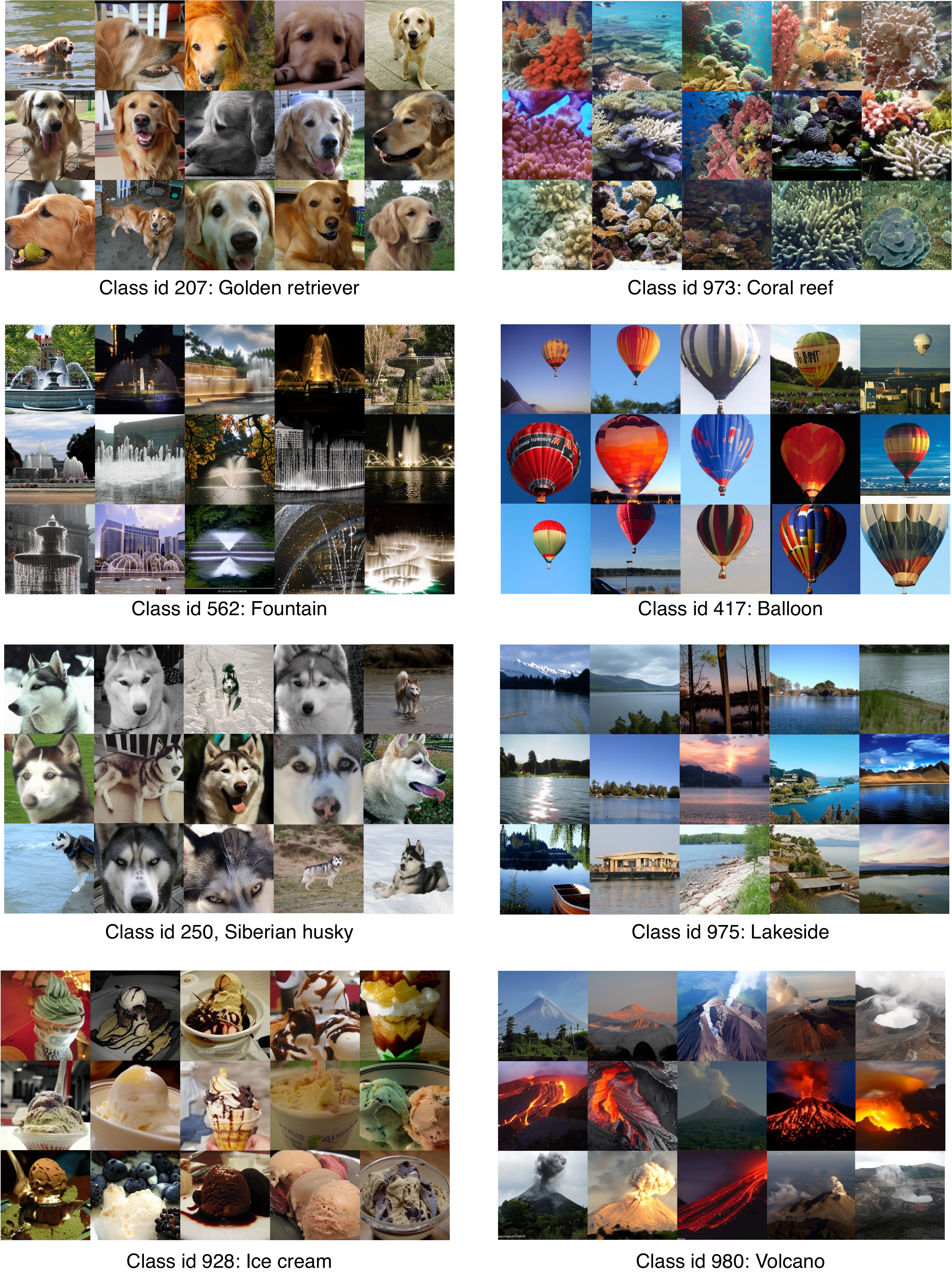}
\vspace{-2mm}
\caption{\textbf{Class-conditional 256$\times$256 Image Generation.} Each block shows samples generated for a specific ImageNet class, labeled with its class id and name. Combined with \ours{}, a vanilla autoregressive generative model achieves strong diversity and fidelity.}
\label{fig:c2i}
\end{figure*}

The class-conditional image generation results are shown in \cref{fig:c2i}. Our autoregressive model achieves a FID of 1.74 on ImageNet, and the qualitative results illustrate its capability to generate visually compelling and diverse images.

\subsection{Adaptive Image Generation} 
\begin{figure*}[!t]
\centering
\includegraphics[width=1.0\textwidth]{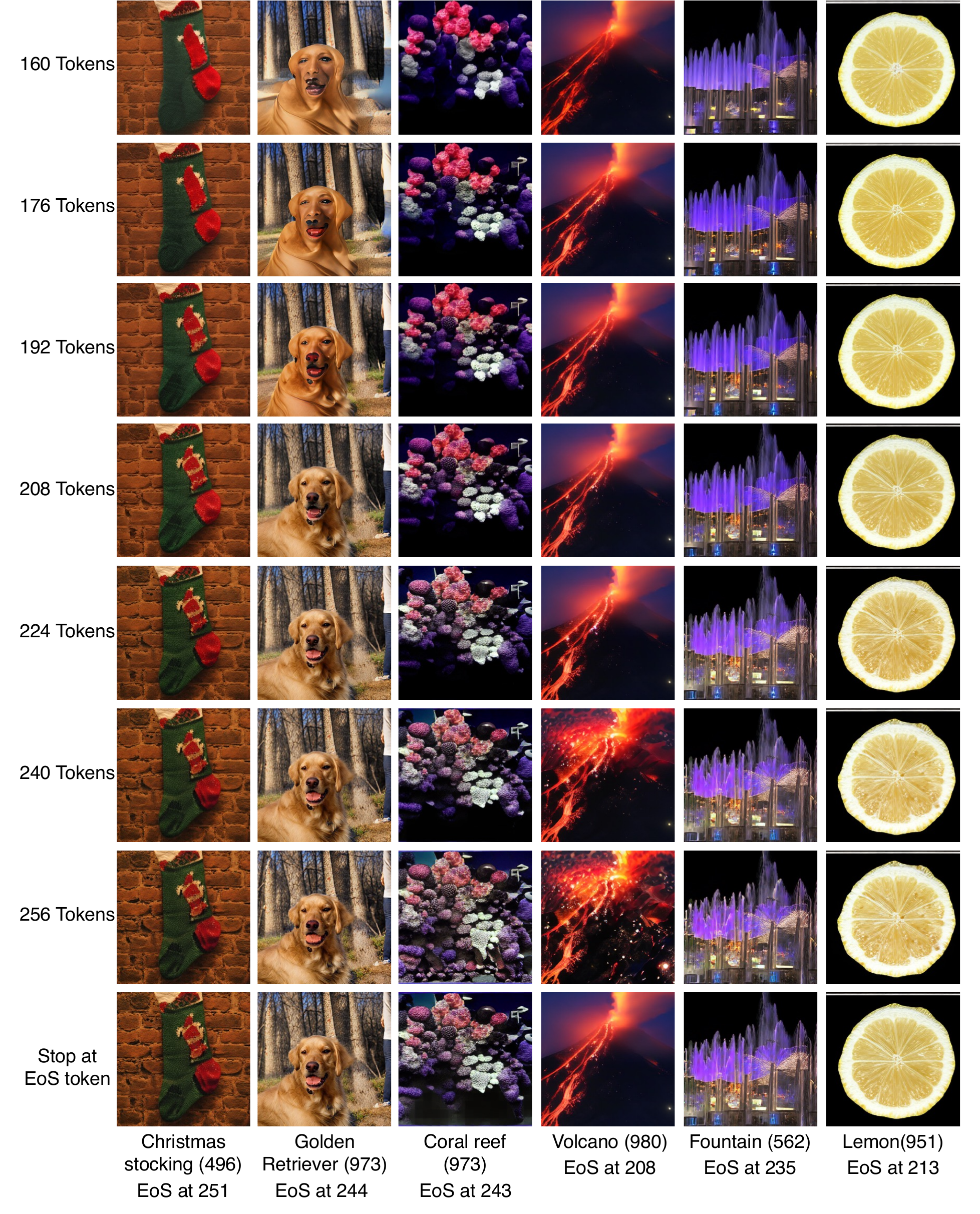}
\vspace{-5mm}
\caption{\textbf{Adaptive Image Generation with \ours{}.} Bottom labels show the class name and the predicted End-of-Sequence (EoS) position for each sample. Rows show images generated with fixed token budgets or the token counts determined by the EoS position.}
\label{fig:adaptive_gen}
\end{figure*}

\cref{fig:adaptive_gen} shows that the adaptivity of \ours{} enables the autoregressive model to perform adaptive image generation as well: the model can generate plausible images with as few as 160 tokens, while allocating more tokens progressively enhances realism and detail. This mirrors the adaptive behavior of \ours{} in image reconstruction.

\section{Limitations}
\label{sec:limitation}
We observe that when the token count used for reconstructing an image deviates significantly from the token count allocated by \ours{}, quality degrades. Using too few tokens naturally leads to poor reconstructions, but using too many tokens can also introduce artifacts and color distortions. Representative examples are provided in \cref{fig:limitation}.

In text-conditional generation, our model still faces challenges such as imperfect text rendering, structural inconsistencies, and mismatches between the generated content and the input prompts. These issues primarily stem from limitations in the scale and quality of available training data, as well as our computational budget. We expect that larger, higher-quality datasets and increased compute will help alleviate these shortcomings.

Finally, in this work we evaluate only the reconstruction capabilities of \ours{} on videos and on images at $512\times512$ resolution. Exploring the generation performance of \ours{} with autoregressive models on videos and at even higher image resolutions would be a direction for future work.

\section{Implementation Details}
\label{sec:implementation}
We provide implementation and training details for \ours{} in \cref{sec:stat_details} and for autoregressive generative models in \cref{sec:ar_details}. All training and inference code, along with model checkpoints, will be released for research purposes.

\subsection{\ours{} Details}
\label{sec:stat_details}
As described in the main paper, \ours{} is trained in two stages. \cref{tables/hyperparameter_stage1} and \cref{tables/hyperparameter_stage2} summarize the full model configurations and training hyperparameters for each stage.

\noindent\textbf{Loss Functions.} 
We use a reconstruction loss $\mathcal{L}_\text{recon} = \mathcal{L}_\text{L2} + \mathcal{L}_\text{perceptual}$, which combines an L2 loss with a perceptual loss to improve visual fidelity. A GAN loss $\mathcal{L}_\text{GAN}$ is included to further enhance reconstruction quality, and a VQ loss $\mathcal{L}_\text{VQ}$ is applied to optimize the codebook. The total loss for the first stage is defined as
\begin{equation}
\mathcal{L}_\text{stage1} = \mathcal{L}_\text{recon} + \mathcal{L}_\text{GAN} + \mathcal{L}_\text{VQ},
\end{equation}
where loss weights are omitted for simplicity. 

In the second stage, we additionally introduce three regularization losses to learn adaptive token allocation, yielding
\begin{equation}
\begin{split}
    \mathcal{L}_\text{stage1} = \mathcal{L}_\text{recon} + \mathcal{L}_\text{GAN} + 
    \mathcal{L}_\text{VQ} \\ + \mathcal{L}_\text{content} + \mathcal{L}_\text{decrease} + \mathcal{L}_\text{sparse},
\end{split}
\end{equation}
again omitting the weighting parameters. Please refer to \cref{tables/hyperparameter_stage1} and \cref{tables/hyperparameter_stage2} for the exact weighting values used in both stages. The loss weights are chosen such that the maximum token count allocated by \ours{} approximately matches the full sequence length (256). For the KL sparsity prior loss $\mathcal{L}_\text{sparse}$, we set the target sparsity to $p^*=0.5$.

\noindent\textbf{Architecture of the Probability Head.} In the second stage, \ours{} employs a probability head to predict per-token keep probabilities. This head is implemented as a position-aware multi-layer perceptron (MLP) with a final sigmoid activation. The MLP consists of a linear layer, a GELU~\cite{hendrycks2016gaussian} activation, and another linear layer. To incorporate positional information, sinusoidal positional embeddings~\cite{vaswani2017attention} are added to the input before it is fed into the MLP.

\begin{table}[t!]
\setlength{\dashlinedash}{0.5pt}
\setlength{\dashlinegap}{1.5pt} 
\setlength{\arrayrulewidth}{0.5pt}
    \centering
    \caption{
    \label{tables/hyperparameter_stage1}
    \textbf{Hyperparameters for the first-stage training of \ours{}.} Most settings follow TiTok~\cite{yu2024image} and TA-TiTok~\cite{kim2025democratizing}, except that our transformer decoder is trained to directly predict pixels rather than discrete codes of MaskGIT VQ-GAN in TiTok.}
    \begin{tabular}{ll}
        \hline
        \textbf{Item} & \textbf{Value} \\
        \hline \hline
        \multicolumn{2}{l}{\textbf{Model}} \\
        \hline
        Codebook Size & 4,096 \\
        Token Size & 12 \\
        Model Size & ViT Large \\
        Patch Size & 16 \\
        Latent Sequence Length & 256 \\
        \hline
        \multicolumn{2}{l}{\textbf{Training}} \\
        \hline
        Epochs & 100 \\
        Batch Size & 1024 \\
        Dataset & ImageNet-1k~\cite{deng2009imagenet} \\
        Augmentation & Random Crop / Flip \\
        \hline
        \multicolumn{2}{l}{\textbf{Losses}} \\
        \hline
        L2 Reconstruction Weight & 1.0 \\
        Quantizer Weight & 1.0 \\
        Perceptual Loss Model & ConvNeXT-Small~\href{https://pytorch.org/vision/main/models/generated/torchvision.models.convnext_small.html#torchvision.models.ConvNeXt_Small_Weights}{[Link]}\\
        Perceptual Loss Weight & 1.1 \\
        Codebook Loss Weight & 1.0 \\
        Commitment Loss Weight & 0.25 \\
        Discriminator Loss Start & Epoch 80 \\
        Discriminator Weight & 0.1 \\
        Lecam Regularization Weight & 0.001 \\
        \hline
        \multicolumn{2}{l}{\textbf{Optimizer}} \\
        \hline
        Optimizer & AdamW~\cite{kingma2014adam} \\
        Learning Rate (LR) & 1e-4 \\
        Discriminator LR & 1e-4 \\
        Beta1 & 0.9 \\
        Beta2 & 0.999 \\
        Weight Decay & 1e-4 \\
        Epsilon & 1e-8 \\
        \hline
        \multicolumn{2}{l}{\textbf{Scheduler}} \\
        \hline
        Scheduler Type & Cosine \\
        Warmup Steps & 10,000 \\
        End Learning Rate & 1e-5 \\
        \hline
    \end{tabular}

\end{table}
\begin{table}[t!]
\setlength{\dashlinedash}{0.5pt}
\setlength{\dashlinegap}{1.5pt} 
\setlength{\arrayrulewidth}{0.5pt}
    \centering
    \caption{
    \label{tables/hyperparameter_stage2}
    \textbf{Hyperparameters for the second stage training of \ours{}}. We use a lower learning rate for the probability head to ensure a more stable training process.}
    \begin{tabular}{ll}
        \hline
        \textbf{Item} & \textbf{Value} \\
        \hline \hline
        \multicolumn{2}{l}{\textbf{Model}} \\
        \hline
        Codebook Size & 4,096 \\
        Token Size & 12 \\
        Model Size & ViT Large \\
        Patch Size & 16 \\
        Latent Sequence Length & 256 \\
        \hline
        \multicolumn{2}{l}{\textbf{Training}} \\
        \hline
        Epochs & 160 \\
        Batch Size & 1024 \\
        Dataset & ImageNet-1k~\cite{deng2009imagenet} \\
        Augmentation & Random Crop / Flip \\
        \hline
        \multicolumn{2}{l}{\textbf{Losses}} \\
        \hline
        L2 Reconstruction Weight & 1.0 \\
        Quantizer Weight & 1.0 \\
        Perceptual Loss Model & ConvNeXT-Small~\href{https://pytorch.org/vision/main/models/generated/torchvision.models.convnext_small.html#torchvision.models.ConvNeXt_Small_Weights}{[Link]}\\
        Perceptual Loss Weight & 1.1 \\
        Codebook Loss Weight & 1.0 \\
        Commitment Loss Weight & 0.25 \\
        Discriminator Loss Start & Epoch 20 \\
        Discriminator Weight & 0.1 \\
        Lecam Regularization Weight & 0.001 \\
        Rank Weight & 1.0 \\
        Decreasing Loss Weight & 50.0 \\
        Sparse Loss Weight & 0.005 \\
        \hline
        \multicolumn{2}{l}{\textbf{Optimizer}} \\
        \hline
        Optimizer & AdamW ~\cite{kingma2014adam}\\
        Learning Rate (LR) & 5e-5 \\
        Probability Head LR & 1e-5 \\
        Discriminator LR & 5e-5 \\
        Beta1 & 0.9 \\
        Beta2 & 0.999 \\
        Weight Decay & 1e-4 \\
        Epsilon & 1e-8 \\
        \hline
        \multicolumn{2}{l}{\textbf{Scheduler}} \\
        \hline
        Scheduler Type & Cosine \\
        Warmup Steps & 10,000 \\
        End Learning Rate & 1e-5 \\
        \hline
    \end{tabular}
\end{table}

\subsection{Autoregressive Model Details}
\label{sec:ar_details}
We train three variants of autoregressive models with sizes XL (775M), XXL (1.4B), and 3B (3.1B). Their architectural configurations follow LlamaGen~\cite{sun2024autoregressive}, except that we replace the 2D RoPE~\cite{su2024roformer} with a 1D version to match the 1D token sequence produced by \ours{}.

\noindent\textbf{Position of the End-of-Sequence (EoS) Token.} 
During training, we dynamically sample a threshold in each forward pass to determine the EoS position. We maintain a set of fixed thresholds $[0.99, 0.5, 0.25, 0.1, 0.01, 0.001]$. With $75\%$ probability, we randomly draw a threshold from this set, and the first token position whose predicted keep probability falls below this threshold is marked as the EoS. For the remaining $25\%$ of iterations, we instead sample a threshold uniformly from $(0,1)$ and determine the EoS position in the same way. Note that this threshold-based EoS sampling is valid only because the keep probability profile is enforced to be non-increasing by the importance decreasing prior used when training \ours{}.

\noindent\textbf{Loss Computation.} The autoregressive model is trained using cross-entropy loss. Once the EoS position is identified, all tokens beyond the EoS are padded and excluded in the computation of the loss, as they provide no meaningful supervision under the chosen decoding length.

\noindent\textbf{Sampling Protocols.} For FID evaluation, we generate 50,000 images and compute the metric with the code from~\cite{dhariwal2021diffusion}. Classifier-free guidance~\cite{ho2022classifier} is applied during sampling. Following RAR~\cite{yu2024randomized}, we adopt the power-cosine guidance schedule~\cite{gao2023masked}. We do not apply any top-\emph{k} or top-\emph{p} filtering techniques.\\

\begin{table}[t!]
\setlength{\dashlinedash}{0.5pt}
\setlength{\dashlinegap}{1.5pt} 
\setlength{\arrayrulewidth}{0.5pt}
    \centering
    \caption{
    \label{tables/hyperparameter_ar}
    \textbf{Hyperparameters for the autoregressive models.}}
    \resizebox{0.475\textwidth}{!}{
    \begin{tabular}{ll}
        \hline
        \textbf{Item} & \textbf{Value} \\
        \hline \hline
        \multicolumn{2}{l}{\textbf{Model}} \\
        \hline
        Layers & 36 (XL) / 48 (XXL) / 24 (3B) \\
        Hidden Size & 1280 (XL) / 1536 (XXL) / 3200 (3B) \\
        Heads & 20 (XL) / 24 (XXL) / 32 (3B) \\
        \hline
        \multicolumn{2}{l}{\textbf{Training}} \\
        \hline
        Epochs & 400 (XL) / 640 (XXL) / 800 (3B) \\
        Batch Size & 512 (XL) / 1024 (XXL) / 1024 (3B) \\
        Dataset & ImageNet-1k~\cite{deng2009imagenet} \\
        Optimizer & AdamW~\cite{kingma2014adam} \\
        Learning Rate & 1e-4 \\
        \hline
        \multicolumn{2}{l}{\textbf{Sampling}} \\
        \hline
        Guidance Schedule & pow-cosine~\cite{gao2023masked} \\
        Temmperature & 1.0 \\
        Scale Power & 2.5 (XL) / 1.2 (XXL) / 1.15 (3B) \\
        Guidance Scale & 18.0 (XL) / 7.0 (XXL) / 12.0 (3B) \\
        \hline
    \end{tabular}
}
\end{table}

\noindent\textbf{Please find all figures on the following pages.}

\end{document}